\documentclass[12pt]{article}

\usepackage{openwork}
\usepackage{authblk}
\usepackage{hyperref}
\usepackage{graphicx}
\usepackage{caption}
\usepackage{placeins}
\usepackage{booktabs}
\usepackage{multirow}
\usepackage{tabularx}
\usepackage{array}
\usepackage{longtable}
\makeatletter
\renewcommand\@makefntext[1]{%
  \noindent\hb@xt@1.8em{\hss\@makefnmark}#1%
}
\makeatother

\makeatletter
\def\ps@openwork{%
  \let\@oddhead\@empty
  \let\@evenhead\@empty
  \def\@oddfoot{\hfil\thepage\hfil}
  \def\@evenfoot{\hfil\thepage\hfil}
}
\def\ps@plain{\ps@openwork}
\def\ps@headings{\ps@openwork}
\def\ps@myheadings{\ps@openwork}
\makeatother

\title{Comp2Comp: Open-Source Software with FDA-Cleared Artificial Intelligence Algorithms for Computed Tomography Image Analysis\\[1em]}

\footnotetext[1]{Shared co-first authorship}
\footnotetext[2]{Shared co-senior authorship}
\author[1]{Adrit Rao\protect\footnotemark[1]}
\author[2,3]{Malte Jensen, PhD\protect\footnotemark[1]}
\author[2,3,4]{Andrea T.\ Fisher, MD\protect\footnotemark[1]}
\author[2,3]{Louis Blankemeier, PhD\protect\footnotemark[1]}
\author[4]{Pauline Berens, MD}
\author[4]{Arash Fereydooni, MD}
\author[5]{Seth Lirette, PhD}
\author[6]{Eren Alkan, BS}
\author[6,7]{Felipe C.\ Kitamura, MD, PhD}
\author[2,3,8]{Juan M.\ Zambrano Chaves, MD, PhD}
\author[2,3]{Eduardo Reis, MD}
\author[3,9,10]{Arjun Desai, PhD}
\author[3]{Marc H.\ Willis, DO, MMM, FACR}
\author[2,11,12]{Jason Hom, MD}
\author[3]{Andrew Johnston, MD, MBA}
\author[3]{Leon Lenchik, MD}
\author[3]{Robert D.\ Boutin, MD}
\author[7,13]{Eduardo M.\ J.\ M.\ Farina, MD}
\author[7,13,14]{Augusto S.\ Serpa, MD}
\author[15]{Marcelo S.\ Takahashi, MD}
\author[16]{Jordan Perchik, MD}
\author[16]{Steven A.\ Rothenberg, MD}
\author[17]{Jamie L.\ Schroeder, MD}
\author[17]{Ross Filice, MD}
\author[18]{Leonardo K.\ Bittencourt, MD, PhD}
\author[19]{Hari Trivedi, MD}
\author[19]{Marly van Assen, PhD}
\author[20]{John Mongan, MD, PhD}
\author[20]{Kimberly Kallianos, MD}
\author[1]{Oliver Aalami, MD\protect\footnotemark[2]}
\author[2,3,21,22,23]{Akshay S.\ Chaudhari, PhD\protect\footnotemark[2]}

\affil[1]{Stanford Mussallem Center for Biodesign, Stanford, CA, USA}
\affil[2]{Stanford Center for Artificial Intelligence in Medicine and Imaging, Stanford, CA, USA}
\affil[3]{Stanford Department of Radiology, Stanford, CA, USA}
\affil[4]{Stanford Division of Vascular Surgery, Department of Surgery, Stanford, CA, USA}
\affil[5]{University of Mississippi Medical Center Department of Data Science, Jackson, MS, USA}
\affil[6]{Bunkerhill Health, San Francisco, CA, USA}
\affil[7]{Departamento de Diagnóstico por Imagem, Universidade Federal de São Paulo, São Paulo, Brazil}
\affil[8]{Microsoft Research, Redmond, WA, USA}
\affil[9]{Stanford Department of Electrical Engineering, Stanford, CA, USA}
\affil[10]{Cartesia AI, San Francisco, CA, USA}
\affil[11]{Stanford Division of Hospital Medicine, Department of Medicine, Stanford, CA, USA}
\affil[12]{Stanford University Human-Centered AI, Stanford, CA, USA}
\affil[13]{Diagnósticos da América S.A., Barueri, São Paulo, Brazil}
\affil[14]{Departamento de Diagnóstico por Imagem, Universidade de São Paulo, São Paulo, Brazil}
\affil[15]{Department of Radiology, University of North Carolina, Chapel Hill, NC, USA}
\affil[16]{Department of Radiology, University of Alabama, Birmingham, AL, USA}
\affil[17]{Department of Radiology, MedStar Georgetown University, Washington, DC, USA}
\affil[18]{Department of Radiology, University Hospitals Cleveland Medical Center, Cleveland, OH, USA}
\affil[19]{Department of Radiology, Emory University, Atlanta, GA, USA}
\affil[20]{Department of Radiology and Biomedical Imaging, University of California San Francisco, San Francisco, CA, USA}
\affil[21]{Stanford Department of Biomedical Data Science, Stanford, CA, USA}
\affil[22]{Stanford Cardiovascular Institute, Stanford, CA, USA}
\affil[23]{Weill Cancer Hub West, Stanford, CA and San Francisco, CA, USA}

\date{}

\begin{document}
\maketitle

\clearpage
\begin{abstract}
\textbf{Background:} Artificial intelligence allows automatic extraction of imaging biomarkers from already-acquired radiologic images. This paradigm of opportunistic imaging adds value to medical imaging without additional imaging costs or patient radiation exposure. However, many open-source image analysis solutions lack rigorous validation while commercial solutions lack transparency, leading to unexpected failures when deployed. Here, we report two of the first fully open-sourced, FDA-510(k)-cleared deep learning pipelines to mitigate both challenges. We describe the development and transparent validation of the Comp2Comp platform for opportunistic analysis of computed tomography scans.
\par
\vspace{0.75em}
\par

\textbf{Methods:} Comp2Comp includes pipelines for Abdominal Aortic Quantification (AAQ) and Bone Mineral Density (BMD) estimation. AAQ segments the abdominal aorta to assess aneurysm size; BMD segments vertebral bodies to estimate trabecular bone density and osteoporosis risk. AAQ-derived maximal aortic diameters were compared against radiologist ground-truth measurements on 258 patient scans enriched for abdominal aortic aneurysms from four external institutions. BMD binary classifications (low vs. normal bone density) were compared against concurrent DXA scan ground truths obtained on 371 patient scans from four external institutions.
\par
\vspace{0.75em}
\par

\textbf{Results:} AAQ had an overall mean absolute error of 1.57 mm (95\% CI 1.38-1.80 mm). BMD had a sensitivity of 81.0\% (95\% CI 74.0-86.8\%) and specificity of 78.4\% (95\% CI 72.3-83.7\%). Both AAQ (K243779) and BMD (K242295) received FDA 510(k) clearance. Towards full transparency, all pivotal data and statistical endpoints submitted to the FDA are included in the manuscript. 
\par
\vspace{0.75em}
\par

\textbf{Conclusions:} Comp2Comp AAQ and BMD demonstrated sufficient accuracy for clinical use. Open-sourcing these algorithms improves transparency of typically opaque FDA clearance processes, allows hospitals to test the algorithms before cumbersome clinical pilots, and provides researchers with best-in-class methods. Comp2Comp has a permissive Apache License 2.0 that allows for use, modification, and distribution for any commercial and non-commercial purposes. Comp2Comp can be found at \url{https://github.com/StanfordMIMI/Comp2Comp}.
\\

\end{abstract}
\clearpage
\section{\textbf{Introduction}}

With 1357 FDA-authorized artificial intelligence (AI) medical devices as of January 2026, AI has recently proliferated in medicine, and radiology comprises the largest category with 1039 entries (77\%).\textsuperscript{1} AI-based image analysis software can assist physicians with a range of imaging interpretation tasks, including quantitative measurement extraction, radiology report generation, and automatic detection of high-risk findings,\textsuperscript{2,3} with some AI models performing all three.\textsuperscript{4} Because computed tomography (CT) is the most common cross-sectional imaging technique in radiology,\textsuperscript{5} many AI solutions focus on CT. Opportunistic CT scan analysis, whereby imaging is reviewed for findings additional to the original indication, has been largely facilitated by AI and holds potential for detection of previously unrecognized pathology such as early-stage cancer or asymptomatic coronary artery disease.\textsuperscript{6-9} 
\par
\vspace{0.75em}
\par
While imaging AI models have continually improving capabilities, clinical adoption and demonstration of value is hindered by high costs and logistical complexity.\textsuperscript{10-13} Prior to our work, all Food and Drug Administration (FDA) cleared AI solutions for opportunistic CT analysis were offered commercially rather than open-source.\textsuperscript{14-16} A “transparency gap” exists for such commercially offered software, whereby clinicians cannot fully evaluate these offerings within their specific workflows and patient populations.\textsuperscript{17,18} Although some commercial solutions may include a low/no-cost trial, code, performance, and clinical validations of closed-source models remain unavailable for full scrutiny by potential adopters.\textsuperscript{19} Hospitals typically develop bespoke frameworks for evaluating such AI models, which divert resources from patient care.\textsuperscript{20,21} Even after careful vetting that requires clinical, financial, legal, and cybersecurity efforts, real-world implementation of AI solutions may yield no clinical benefit.\textsuperscript{22} Open-source models have addressed these challenges and considerably improve adoption speed by providing the transparency necessary for local validation and reducing the administrative burden on healthcare systems.\textsuperscript{16}
\par
\vspace{0.75em}
\par
Software as medical devices that are cleared by the FDA have publicly available summary statements describing the pivotal studies that were performed to secure FDA clearance. However, device manufacturers are not required to share all specific study details in summary statements, creating an opaqueness in how medical devices are cleared.\textsuperscript{18,23} Making this process fully transparent may help predict success in local deployments, engender broader trust in the devices, and accelerate future research. 
\par
\vspace{0.75em}
\par
To address this transparency limitation of closed-source AI algorithms, we present the Computed Tomography to Body Composition (Comp2Comp) platform. We describe pivotal clinical trial results for two FDA 510(k) cleared modules for opportunistic body composition analysis, Abdominal Aortic Quantification (AAQ) and Bone Mineral Density (BMD), whose code and model weights are open sourced within Comp2Comp. AAQ automatically measures the maximum diameter of the abdominal aorta to opportunistically detect abdominal aortic aneurysms (AAA). Aortic aneurysms account for 172,000 annual deaths globally and are catastrophic if not identified and treated prior to rupture, making early identification paramount.\textsuperscript{24} BMD automatically estimates lumbar vertebral body trabecular bone density\textsuperscript{25} to opportunistically identify patients with low bone mineral density (osteopenia). The high prevalence of osteoporosis leads to a high burden of disability from osteoporotic fractures,\textsuperscript{26} and given the low rates of dual-energy X-ray absorptiometry (DXA) screening, opportunistic detection on CT could provide more patients with preventive treatment options.\textsuperscript{27,28} 
\par
\vspace{0.75em}
\par
In this manuscript, we report all pivotal data and statistical analysis plans exactly as they were included in the FDA 510(k) filings to maximize transparency. By making the Comp2Comp software publicly available and publishing the details of our pipeline and validation, we hope to provide prospective adopters with all necessary information to determine if trial or full adoption is worthwhile. Moreover, Comp2Comp can support researchers globally with open access to best-in-class FDA-cleared methods and provide transparency on an otherwise opaque FDA clearance process for current and future AI-enabled medical devices. The Comp2Comp repository with AAQ and BMD modules can be found here: \href{https://github.com/StanfordMIMI/Comp2Comp}{https://github.com/StanfordMIMI/Comp2Comp}

\subsection{\textbf{\textbf{Results}}}
\textit{Approval and Indications for Use}
\par
\vspace{0.75em}
\par
Both the AAQ and BMD models validated in this study are available within the Comp2Comp open-source software package. AAQ received FDA 510(k) clearance on July 1, 2025 while BMD received clearance on April 8, 2025, with device numbers K243779 and K242295, respectively. Intended use of each model is as follows: AAQ post-processing AI-powered software analyzes the abdominal aorta in the L1-L5 lumbar vertebral region and produces the maximum axial abdominal aortic diameter in adults aged 18 and older who have not yet undergone abdominal aortic surgery. BMD post-processing AI-powered software estimates DXA-measured average area bone mineral density of vertebral bodies from existing CT scans in adults 30 years and above, and it outputs a flag for low bone density below a pre-specified threshold but does not replace DXA or any other test dedicated to BMD measurement. 
\par
\vspace{0.75em}
\par
We note the subtle, yet important, distinction that the Comp2Comp platform itself does not have FDA clearance. Instead, the AAQ and BMD modules available on Comp2Comp were passed without changes through the quality management system maintained by Bunkerhill Health for eventual clearance and marketing. 
\par
\vspace{1em}
\par
\textit{Automated Pipeline Structures and Outputs}
\par
\vspace{0.75em}
\par
The AAQ pipeline uses a nnU-Net segmentation model\textsuperscript{29} to automatically segment the abdominal aorta for outputting the maximal aortic diameter. Several secondary quality-control outputs are also generated including the slice index of maximal diameter, a graph of diameter throughout the aorta, and a video of the segmentations, but these outputs are not assessed in the pivotal trial. Figure 1 demonstrates AAQ pipeline architecture, training and evaluation datasets, and algorithm primary and secondary outputs. Examples were chosen to demonstrate performance in contrast and non-contrast enhanced scans among aneurysmal, thrombus-laden aortas. 
\par
\vspace{0.75em}
\par
The BMD pipeline also uses a custom nnU-Net model to segment the L1-L4 lumbar vertebrae from an abdominopelvic CT to automatically estimate vertebral bone mineral density from vertebral trabecular bone regions of interest (ROI). The pipeline normalizes the radiodensity (measured via CT Hounsfield Units [HU]) of the ROI to that of visceral adipose tissue and air, resulting in an estimated CT BMD score. A binary classifier compares this CT BMD score to a threshold corresponding to a dual-energy X-ray absorptiometry (DXA) scan T-score threshold of -1.0, to return an output of normal versus low BMD (osteopenia). 
\par
\vspace{0.75em}
\par
\begin{figure}[htbp]
  \centering
  \includegraphics[width=0.8\linewidth]{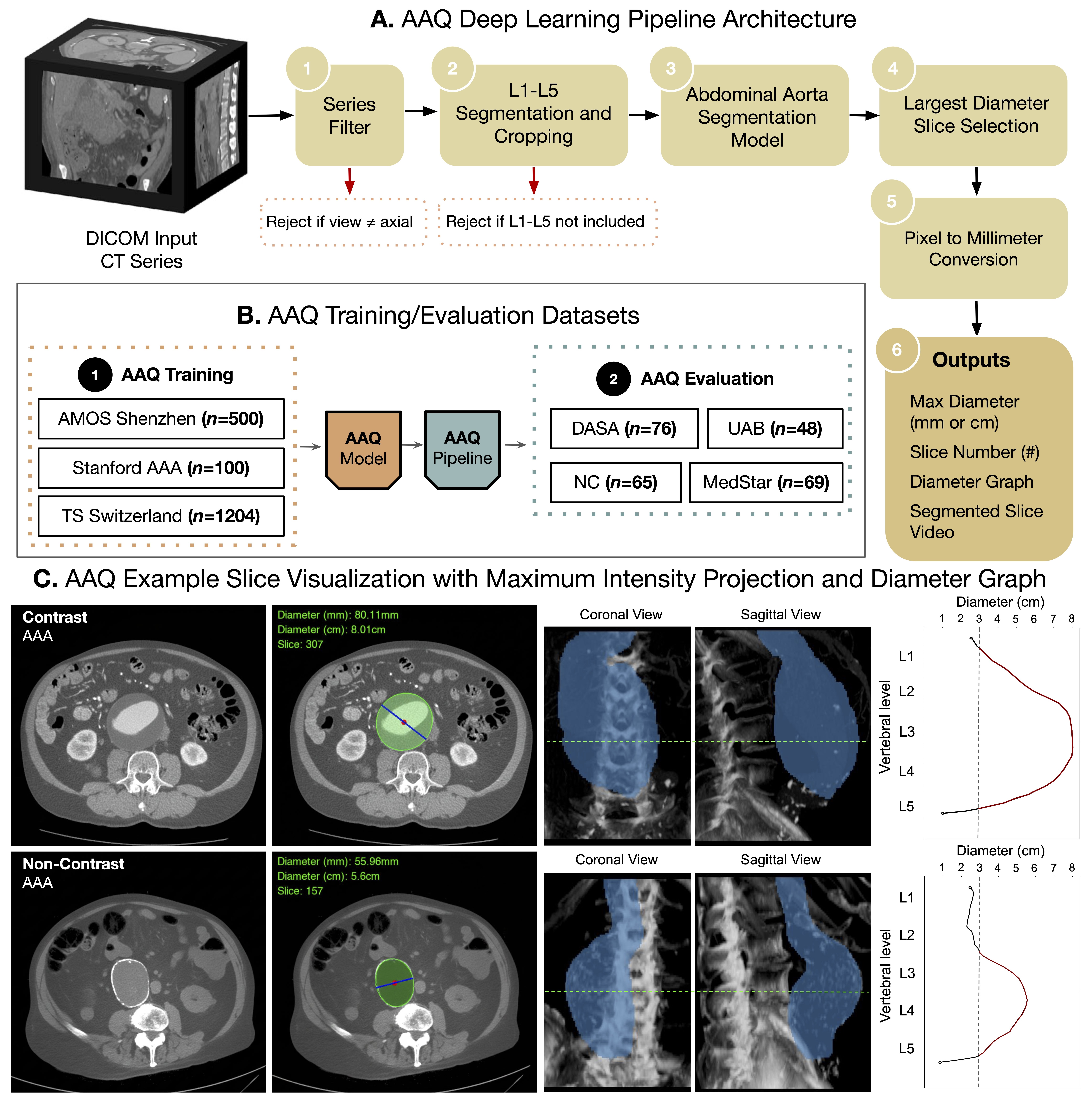}
  \caption{\textbf{(A) }Abdominal aortic quantification (AAQ) pipeline architecture from input DICOM files of CT scan to output of aortic maximum diameter with associated statistics and visualizations.\textbf{ (B) }Study layout, consisting of data acquisition for AAQ training, integration within the AAQ pipeline, and evaluation across multiple institutions. (\textbf{C}) Representative AAQ outputs from contrast enhanced and non-contrast enhanced CT scans. The red dot denotes the center of the segmentation slice, and the blue line denotes the reported diameter measurement. The dashed vertical line on the graph is at 3 cm, the threshold size for aortic aneurysm. \textbf{(D)} Example maximum intensity projection of an abdominal aortic aneurysm (not part of algorithm output).\\ 
  TS = TotalSegmentator\textsuperscript{30}\\
  DASA = Diagnósticos da América S.A.\\
  UAB = University of Alabama at Birmingham\\
  NC = North Carolina\\
  MedStar = MedStar Health}
\end{figure}
\FloatBarrier

\begin{figure}[htbp]
  \centering
  \includegraphics[width=0.8\linewidth]{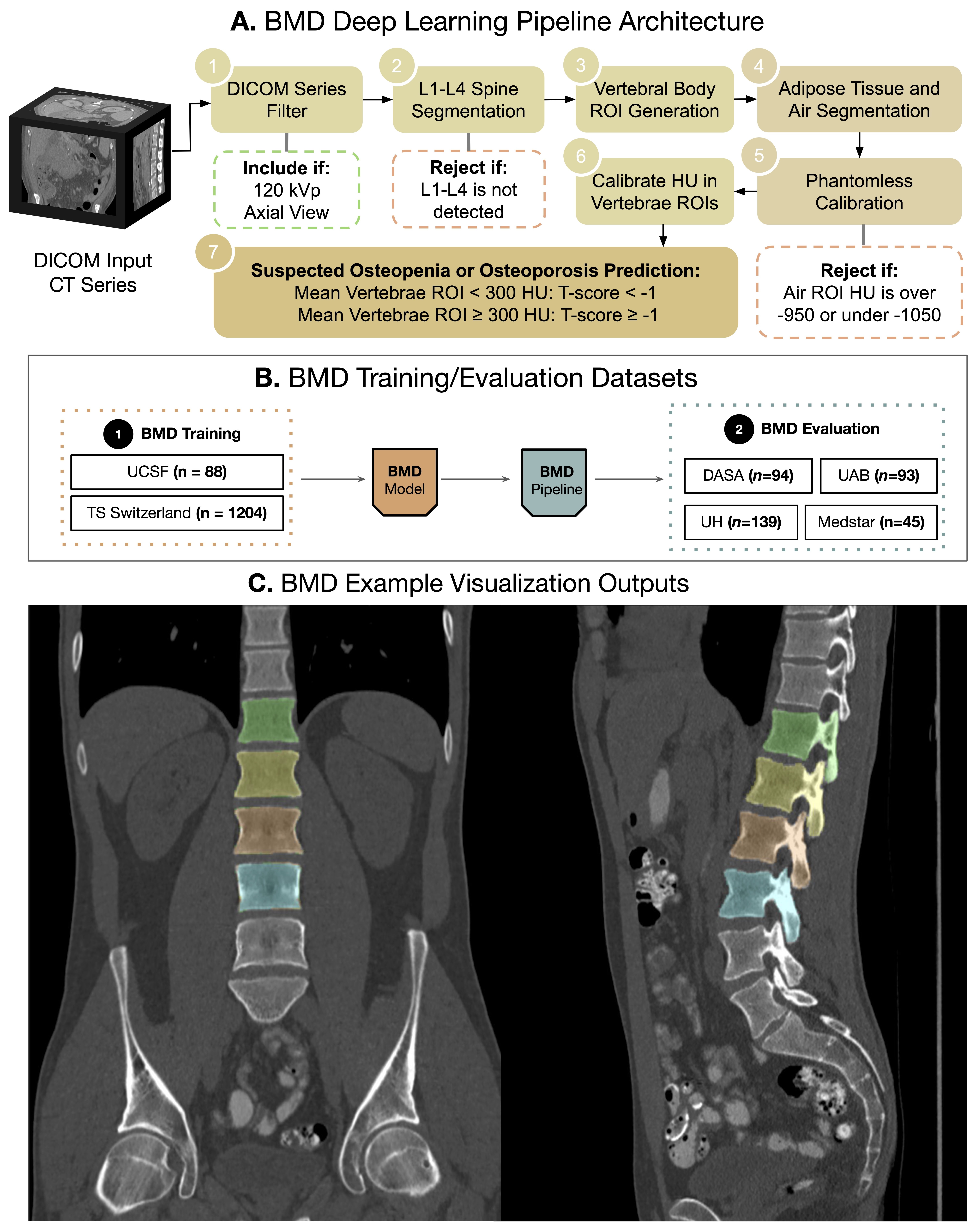}
  \caption{Overview of the BMD Algorithm Development and Output. \textbf{(A)} Processing pipeline from DICOM series input to BMD classification (normal or low density). The pipeline segments the L1-L4 vertebrae and generates ROIs in the body of the vertebrae. The Hounsfield Unit (HU) values are then calibrated by segmented visceral adipose tissue and a ROI placed in air. The final output is a binary prediction of an estimated DXA T-score $\ge -1$ (normal) or $< -1$ (osteopenic). \textbf{(B)} Development and FDA approval cohorts, showing distributions by sex, imaging center, age, and CT scanner vendor. \textbf{(C) }Algorithm output displays coronal and sagittal views with overlaid spine segmentations and ROIs, including median HU values for both the complete vertebra and the ROI.\\
  VerSe = Vertebrae labeling and segmentation benchmark dataset\textsuperscript{31}\\
UCSF = University of California San Francisco
TS = TotalSegmentator\textsuperscript{30}\\
DASA = Diagnósticos da América S.A.\\
UAB = University of Alabama at Birmingham\\
UH = University Hospitals Cleveland Medical Center\\
MedStar = MedStar Health}
\end{figure}
\FloatBarrier

 \textit{Abdominal Aortic Quantification Validation}
\par
\vspace{0.75em}
\par
AAQ’s performance was validated in a curated testing dataset by comparing each maximal aortic diameter output against the average of three board-certified radiologist ground-truth measurements taken from the same scan.  The validation dataset consisted of 258 CT scans from 258 unique patients collected from four institutions with a wide breadth of scanners, patient demographics, and aortic morphology (Table 1). An aneurysm was present in 94 patients (36\%). 
\par
\vspace{0.75em}
\par
Female patients comprised the slight majority (140/258, 54\%), and a plurality of patients were in the 60-79 age group (124/258, 48\%). CT scans performed on Toshiba and Siemens machines were slightly more common (83/258, 32\% and 79/258, 31\%, respectively) compared to those performed on GE and Philips machines (54/258, 21\% and 42/258, 16\%, respectively). Lower slice thickness scans were more common (117/258 $<$ 2.5 mm, 45\%). The majority of scans were completed at 120 kVp (183/258, 71\%), with approximately equal split between contrast enhanced and non-contrast enhanced scans. Scans were split evenly between sites.

\begin{table}[htbp]
\centering
\small
\setlength{\tabcolsep}{6pt}
\renewcommand{\arraystretch}{0.95}

\begin{tabular}{p{0.3\linewidth}p{0.36\linewidth}r r}
\toprule
\textbf{Patient/scan information} &  & \textbf{Count (n=258)} & \textbf{Percentage} \\
\midrule

\textbf{Sex} & Male   & 118 & 46\% \\
            & Female & 140 & 54\% \\
\midrule

\textbf{Age} & 18--39 & 24  & 9\%  \\
            & 40--59 & 68  & 26\% \\
            & 60--79 & 124 & 48\% \\
            & 80+    & 42  & 16\% \\
\midrule

\textbf{CT Manufacturer} & Toshiba & 83 & 32\% \\
                        & Siemens & 79 & 31\% \\
                        & Philips & 42 & 15\% \\
                        & GE      & 54 & 21\% \\
\midrule

\textbf{Slice Thickness} & $<2.5$ mm    & 117 & 45\% \\
                        & 2.5--3.9 mm  & 81  & 31\% \\
                        & $\ge 4.0$ mm & 60  & 23\% \\
\midrule

\textbf{Kilovoltage Peak (kVp)} & $<100$   & 15  & 6\%  \\
                               & 100--119 & 43  & 17\% \\
                               & $=120$   & 183 & 71\% \\
                               & $>120$   & 17  & 7\%  \\
\midrule

\textbf{Contrast Use} & Non-Contrast & 134 & 52\% \\
                     & Contrast     & 124 & 48\% \\
\midrule

\textbf{Site} & Diagnósticos da América S.A.        & 76 & 29\% \\
             & North Carolina                        & 65 & 25\% \\
             & University of Alabama at Birmingham   & 48 & 19\% \\
             & MedStar Health                        & 69 & 27\% \\
\midrule

\textbf{Aneurysm (aorta $>$3 cm)}& No aneurysm & 164 & 64\% \\
                                 & Aneurysm    & 94  & 36\% \\
\bottomrule
\end{tabular}

\caption{Patient demographics, scan characteristics, collection site, and aneurysm status for the AAQ analysis.}
\label{tab:patient-scan}
\end{table}
Mean absolute error (MAE) for the overall dataset was 1.578 mm (95\% CI 1.375-1.797 mm), within the pre-determined 2.0 mm MAE threshold set for acceptable standalone model performance endpoints. Figure 1C demonstrates sample AAQ outputs with overlaid segmentation masks to allow verification of accurate measurement.
\par
\vspace{0.75em}
\par
The intraclass correlation coefficient confidence interval (ICC CI) was used to determine how closely the AAQ measurement clustered with measurements from 3 radiologists. The ICC of radiologists compared to the model measurements was 0.983 (95\% CI 0.977-0.986) compared with 0.985 ICC (95\% CI 0.979-0.989) for inter-reader radiologist variability. The upper limit of difference between confidence intervals was 0.003, well within the pre-determined threshold of 0.05 ICC CI. 
\par
\vspace{1em}
\par
\textit{AAQ Subgroup Analyses}
\par
\vspace{0.75em}
\par

MAE and ICC were calculated for all subgroups, as demonstrated in Table 2. Algorithm performance was equivalent across sex, CT manufacturer, slice thickness, and site. Differences were observed in the other four categories: AAQ performance was significantly improved in patients aged 18-39 (MAE: 0.777 mm in 18-39 years vs. 1.670 mm in 40+ years, p = 0.02). Model performance was significantly worse for scans with kVp $<$ 100 (MAE: 3.086 mm in kVp $<$100 vs. 1.484 mm in kVp $\ge$ 100, p = 0.0004. CT contrast use also affected accuracy, with higher accuracy in scans without intravenous contrast (MAE 1.293 mm in non-contrast vs. 1.885 mm in contrast-enhanced scans, p = 0.005). Using 3 cm as the threshold for abdominal aortic aneurysm, model accuracy was compared for aneurysmal (94 out of 258) vs. nonaneurysmal aortas. Model accuracy was significantly lower for scans with aneurysm compared to those without (aneurysm MAE = 2.220 mm vs. no aneurysm MAE = 1.210 mm, p $<$ 0.0001), potentially related to wall contour irregularities present with aneurysms.

\begingroup
\footnotesize
\setlength{\tabcolsep}{3pt}
\renewcommand{\arraystretch}{0.90}

\setlength{\LTleft}{0pt}
\setlength{\LTright}{0pt}
\setlength{\LTpost}{-6pt}

\begin{longtable}{
>{\raggedright\arraybackslash}p{0.26\linewidth}  % Subgroup Type
l                                                % Count
>{\raggedright\arraybackslash}p{0.20\linewidth}  % MAE
l                                                % R-R ICC
l                                                % R-M ICC
>{\raggedright\arraybackslash}p{0.20\linewidth}  % ICC diff
}

\toprule
\textbf{Subgroup Type} & \textbf{Count} & \textbf{MAE (95\% CI) (mm)} &
\textbf{R--R ICC} & \textbf{R--M ICC} & \textbf{ICC diff (95\% CI)} \\
\midrule
\endfirsthead

\toprule
\textbf{Subgroup Type} & \textbf{Count} & \textbf{MAE (95\% CI) (mm)} &
\textbf{R--R ICC} & \textbf{R--M ICC} & \textbf{ICC diff (95\% CI)} \\
\midrule
\endhead

\midrule
\multicolumn{6}{r}{\emph{Continued on next page}}\\
\endfoot

\bottomrule
\addlinespace[6pt]
\caption{Testing data distribution across relevant variables with associated subgroup MAE, Radiologist--Radiologist ICC (R--R ICC), Radiologist--Model ICC (R--M ICC), and ICC CI difference.\\
\emph{* Significant difference present between the identified category and the remainder of its subgroup.}}
\label{tab:subgroup-performance}
\endlastfoot

Overall & 258 & 1.578 (1.375--1.797) & 0.985 & 0.983 & 0.002--0.003 \\
\midrule

\multicolumn{6}{c}{\textbf{Subgroup Analysis by Sex}} \\
\cmidrule(lr){1-6}
Male   & 118 & 1.537 (1.300, 1.800) & 0.991 & 0.989 & (0.002, 0.002) \\
Female & 140 & 1.611 (1.308, 1.956) & 0.972 & 0.966 & (0.003, 0.006) \\
\midrule

\multicolumn{6}{c}{\textbf{Subgroup Analysis by Age}} \\
\cmidrule(lr){1-6}
18--39* & 24  & 0.777 (0.530, 1.033) & 0.812 & 0.828 & (-0.080, -0.000) \\
40--59  & 68  & 1.215 (0.906, 1.650) & 0.957 & 0.954 & (-0.002, 0.003) \\
60--79  & 124 & 1.841 (1.523, 2.197) & 0.987 & 0.982 & (0.004, 0.006) \\
80+     & 42  & 1.845 (1.422, 2.319) & 0.978 & 0.977 & (-0.002, 0.001) \\
\midrule

\multicolumn{6}{c}{\textbf{Subgroup Analysis by CT Manufacturer}} \\
\cmidrule(lr){1-6}
Toshiba & 83 & 1.823 (1.391, 2.312) & 0.980 & 0.971 & (0.007, 0.010) \\
Siemens & 79 & 1.469 (1.140, 1.871) & 0.981 & 0.980 & (-0.001, 0.002) \\
Philips & 42 & 1.430 (1.049, 1.883) & 0.983 & 0.982 & (-0.003, 0.001) \\
GE      & 54 & 1.473 (1.157, 1.825) & 0.992 & 0.990 & (0.001, 0.003) \\
\midrule

\multicolumn{6}{c}{\textbf{Subgroup Analysis by Slice Thickness}} \\
\cmidrule(lr){1-6}
$<2.5$ mm    & 117 & 1.716 (1.401, 2.075) & 0.983 & 0.980 & (0.001, 0.003) \\
2.5--3.9 mm  & 81  & 1.580 (1.291, 1.906) & 0.988 & 0.986 & (0.000, 0.002) \\
$\ge 4.0$ mm & 60  & 1.304 (0.929, 1.795) & 0.986 & 0.976 & (0.006, 0.014) \\
\midrule

\multicolumn{6}{c}{\textbf{Subgroup Analysis by Kilovoltage Peak (kVp)}} \\
\cmidrule(lr){1-6}
$<100$*  & 15  & 3.086 (1.724, 4.714) & 0.984 & 0.973 & (0.004, 0.029) \\
100--119 & 43  & 1.785 (1.270, 2.395) & 0.977 & 0.972 & (0.003, 0.006) \\
$=120$   & 183 & 1.431 (1.227, 1.655) & 0.987 & 0.986 & (0.000, 0.002) \\
$>120$   & 17  & 1.304 (0.800, 1.944) & 0.917 & 0.917 & (-0.011, 0.001) \\
\midrule

\multicolumn{6}{c}{\textbf{Subgroup Analysis by Contrast Use}} \\
\cmidrule(lr){1-6}
Non-Contrast* & 134 & 1.293 (1.082, 1.543) & 0.979 & 0.978 & (-0.001, 0.002) \\
Contrast      & 124 & 1.885 (1.550, 2.257) & 0.987 & 0.983 & (0.003, 0.006) \\
\midrule

\multicolumn{6}{c}{\textbf{Subgroup Analysis Across Sites}} \\
\cmidrule(lr){1-6}
Diagnósticos da América S.A. & 76 & 1.860 (1.406, 2.386) & 0.981 & 0.972 & (0.007, 0.010) \\
North Carolina               & 65 & 1.499 (1.129, 1.960) & 0.983 & 0.981 & (0.001, 0.001) \\
University of Alabama at Birmingham & 48 & 1.718 (1.350, 2.125) & 0.980 & 0.976 & (0.002, 0.011) \\
MedStar Health               & 69 & 1.243 (0.974, 1.549) & 0.965 & 0.965 & (-0.008, 0.002) \\
\midrule

\multicolumn{6}{c}{\textbf{Subgroup Analysis by Aneurysm Status (Aorta $>3$ cm)}} \\
\cmidrule(lr){1-6}
No aneurysm* & 164 & 1.210 (0.999, 1.420) & 0.843 & 0.836 & (-0.020, 0.015) \\
Aneurysm     & 94  & 2.220 (1.807, 2.632) & 0.972 & 0.963 & (-0.021, 0.005) \\

\end{longtable}
\endgroup
A Bland-Altman plot was generated (Figure 3) demonstrating a mean difference of 0.50 mm between ground truth and AAQ output, indicating that AAQ on average slightly underestimates aortic size. Most of the data points fell within the accepted 5 mm limit of agreement (Figure 3).\textsuperscript{32} The scatter plot of AAQ output vs. ground truth measurement demonstrates reasonable concurrence between true aortic diameter and model estimate at a variety of aortic sizes (Pearson r = 0.982, 95\% CI 0.977-0.986).

\begin{figure}[htbp]
  \centering
  \includegraphics[width=0.85\linewidth]{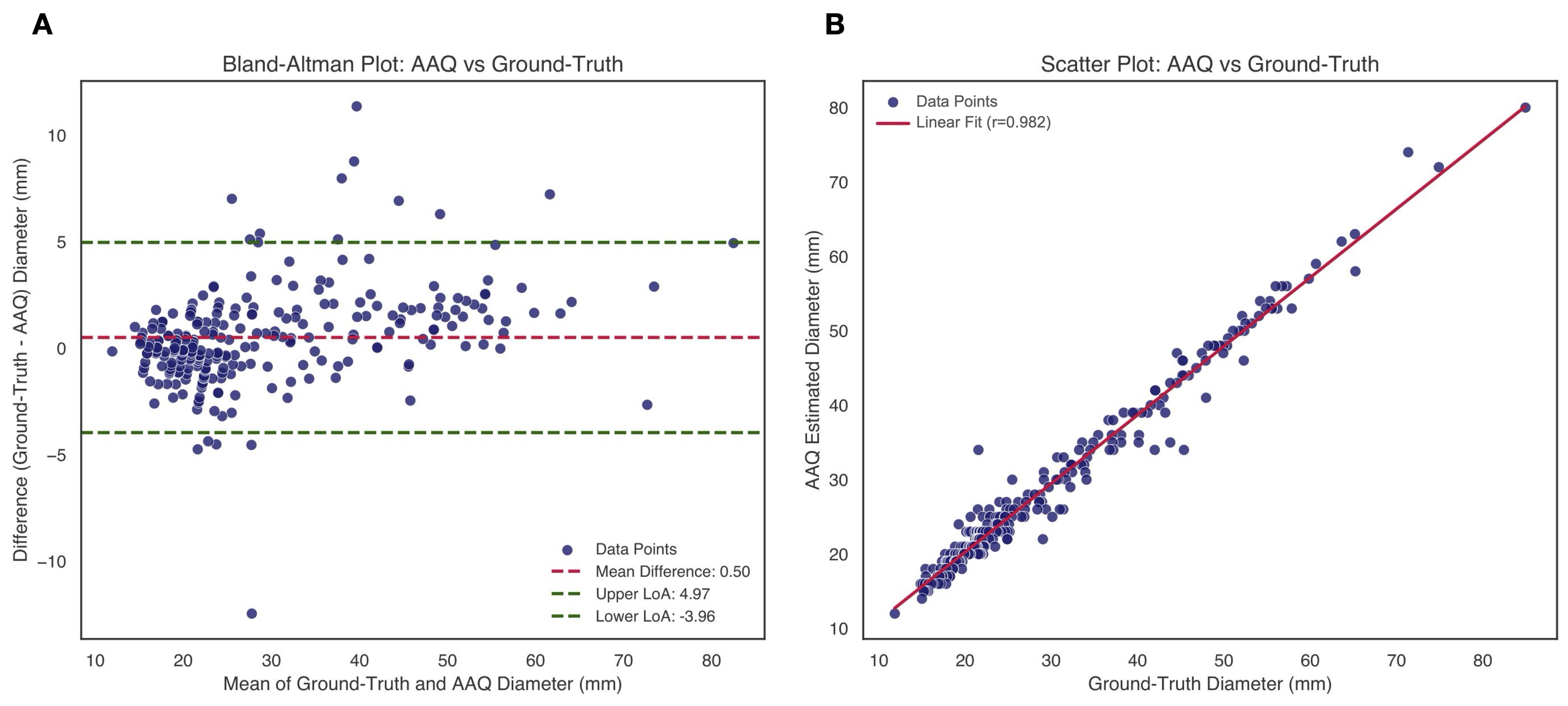}
  \caption{\textbf{(A)} Bland-Altman plot demonstrating relationship between algorithm accuracy and aortic diameter. \textbf{(B)} Scatter plot demonstrating close linear relationship between ground truth and AAQ measurement of aortic diameter. LOA = Limit of Agreement}
\end{figure}
\FloatBarrier

 The model was also evaluated on 39 scans from patients with aortic endografts, who were therefore postoperative from endovascular intervention for abdominal aortic aneurysm. This group was excluded from primary analysis and from the application for FDA clearance.  For this set, accuracy was significantly lower compared to scans acquired from patients without endografts (MAE = 3.964 mm, 95\% CI 2.394-5.534 mm, p $<$ 0.0001). Inter-radiologist ICC was 0.877 and radiologist vs. model ICC was 0.848 with ICC CI difference of 0.019-0.034.
\par
\vspace{1em}
\par
\textit{Bone Mineral Density Validation}
\par
\vspace{0.75em}
\par
BMD model performance was evaluated in a testing dataset of 371 CT scans from 371 unique patients collected from four institutions with a representative distribution of scanners and patient characteristics. Most patients were female (308/371, 83\%), and the majority were in the 50–69-year age range (211/371, 57\%), reflecting the demographics of patients who most commonly received DXA scans at the participating institutions. CT scans performed on GE and Philips scanners were more common (122/371, 33\% and 151/371, 41\%, respectively) than those performed on Siemens and Toshiba scanners (66/371, 18\% and 32/371, 9\%, respectively). All GE scans were reconstructed using Standard kernel, and most Philips scans were reconstructed with B kernel (131/151, 87\%). Slice thickness of 3.0 mm was most common (169/371, 46\%). University Hospitals Cleveland Medical Center (UH) contributed the most patients (139/371, 37\%). 158 patients in the test set of 371 CT/DXA pairs had a ground truth DXA T-score showing low bone mineral density (43\% prevalence) (Table 3).

\begingroup
\centering
\small
\setlength{\tabcolsep}{6pt}
\renewcommand{\arraystretch}{0.95}

\begin{longtable}{p{0.28\linewidth}p{0.37\linewidth}r r}
\toprule
\textbf{Patient/scan information} &  & \textbf{Count (n = 371)} & \textbf{Percentage} \\
\midrule
\endfirsthead

\toprule
\textbf{Patient/scan information} &  & \textbf{Count (n = 371)} & \textbf{Percentage} \\
\midrule
\endhead

\midrule
\multicolumn{4}{r}{\emph{Continued on next page}} \\
\endfoot

\bottomrule
\addlinespace[6pt]
\captionsetup{justification=raggedright,singlelinecheck=false}
\caption{Patient and scan characteristics for the BMD test cohort (n = 371).}
\label{tab:bmd-cohort}
\endlastfoot

\textbf{Sex} & Male   & 63  & 17\% \\
            & Female & 308 & 83\% \\
\midrule

\textbf{Age} & 18--21 & 1   & $<1$\% \\
            & 22--29 & 5   & 1\% \\
            & 30--39 & 15  & 4\% \\
            & 40--49 & 29  & 8\% \\
            & 50--69 & 211 & 57\% \\
            & 70+    & 110 & 30\% \\
\midrule

\textbf{CT Manufacturer} & GE      & 122 & 33\% \\
                        & Philips & 151 & 41\% \\
                        & Siemens & 66  & 18\% \\
                        & Toshiba & 32  & 9\% \\
\midrule

\textbf{Kernel} & GE -- STANDARD   & 122 & 33\% \\
               & Philips -- B     & 131 & 35\% \\
               & Philips -- Other & 20  & 5\% \\
               & Siemens -- B30f  & 21  & 6\% \\
               & Siemens -- Other & 45  & 12\% \\
               & Toshiba -- FC08  & 19  & 5\% \\
               & Toshiba -- Other & 13  & 4\% \\
\midrule

\textbf{Slice Thickness} & $<2.0$ mm   & 77  & 21\% \\
                        & 2.0--2.5 mm & 68  & 18\% \\
                        & 3.0 mm      & 169 & 46\% \\
                        & $>3.0$ mm   & 57  & 15\% \\
\midrule

\textbf{Site} & MedStar Health                      & 45  & 12\% \\
             & Diagnósticos da América S.A.        & 94  & 25\% \\
             & University Hospitals Cleveland      & 139 & 37\% \\
             & University of Alabama at Birmingham & 93  & 25\% \\
\midrule

\textbf{Ground truth bone mineral density} & Normal & 213 & 57\% \\
                                          & Low    & 158 & 43\% \\

\end{longtable}
\endgroup

 In contrast with the AAQ validation process, the BMD model outputs (normal vs. low vertebral body bone mineral density) were compared against DXA scan T-scores from the same patients within a 6-month window of the CT scan. The DXA scan T-score was chosen as the ground truth since this test is the accepted gold standard for diagnosing osteoporosis and osteopenia.\textsuperscript{33} For the primary endpoint, a DXA scan T-score threshold greater or less than -1.0 was compared to each patient’s BMD model binary classification, where a T-score $<$ -1.0 indicated low BMD indicative of osteopenia. Model sensitivity was calculated to be 81.0\% (95\% CI 74.0-86.8\%) and specificity was 78.4\% (95\% CI 72.3-83.7\%), above the pre-determined 70\% threshold established for acceptable model performance. Positive predictive value was found to be 73.6\% (95\% CI 66.4\%-79.9\%), and negative predictive value was 84.8\% (95\% CI 79.0\%-89.5\%). The occurrences of true positives, true negatives, false positives, and false negatives were 128 (35\%), 167 (45\%), 46 (12\%), and 30 (8\%) respectively, when comparing the BMD output to the DXA reference.
\par
\vspace{0.75em}
\par
Because the model generates continuous, non-binary CT BMD scores as an intermediate step in the pipeline, Pearson correlation coefficient between model-generated continuous BMD score and DXA T-score was computed as a secondary endpoint. We found Pearson r = 0.791 (95\% CI 0.752-0.830), which surpassed the pre-determined acceptance criteria set at $>$ 0.70. The associated Bland-Altman and scatter plots are depicted in Figure 4A and 4B.
\par
\vspace{0.75em}
\par
 Area under receiver-operator curve (AUROC) was also computed to compare continuous BMD-predicted T-scores against ground truth DXA T-scores, and to compare the BMD binary predictions of whether T-score would be greater or less than -1 against the DXA-obtained ground truth classifications. Continuous BMD prediction AUROC was 0.883 (95\% CI 0.849-0.916) and binary BMD prediction AUROC was 0.797 (95\% CI 0.756-0.838) (Figure 4C and 4D).
 
\begin{figure}[htbp]
  \centering
  \includegraphics[width=\linewidth]{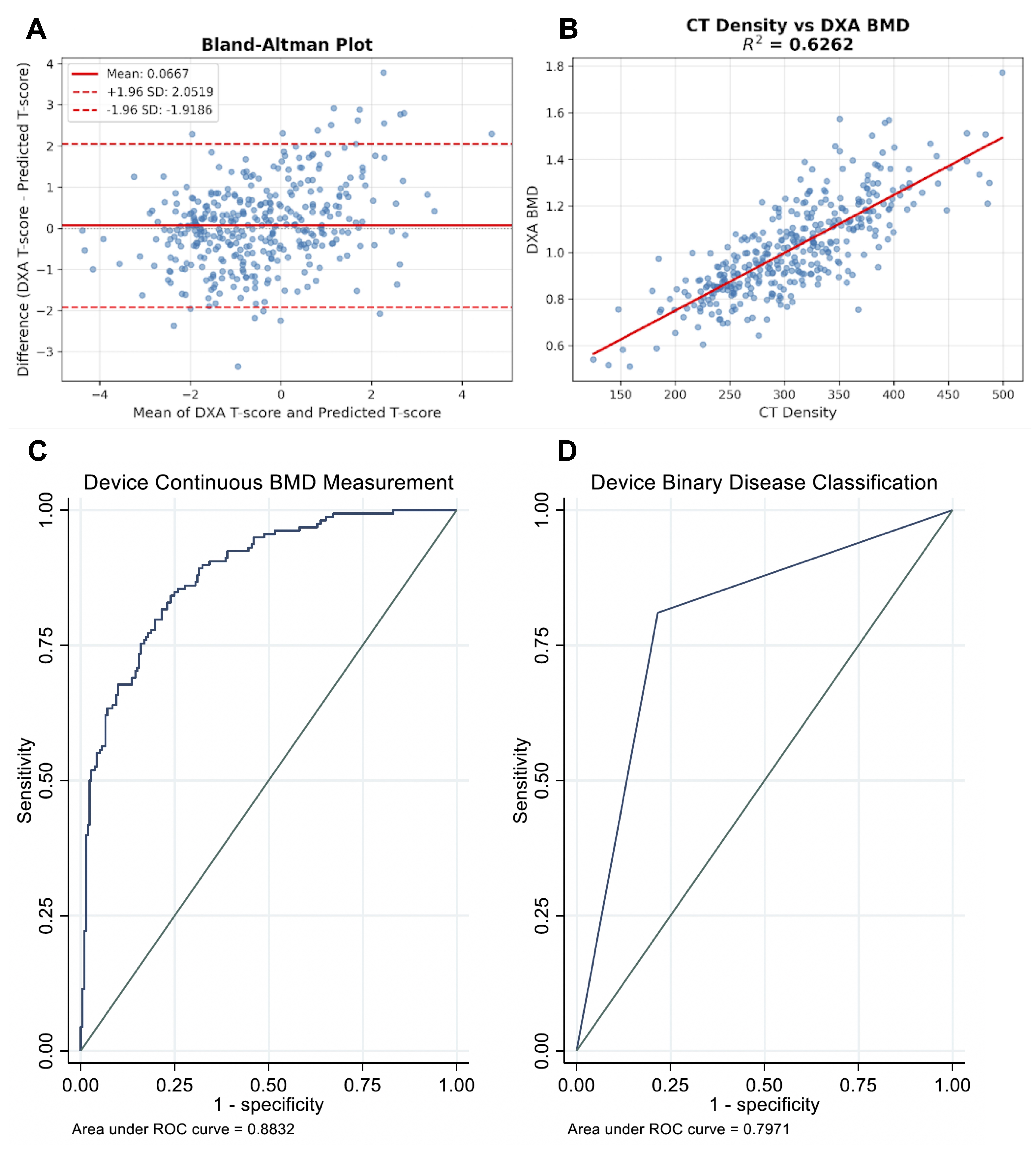}
  \caption{\textbf{(A)} Bland-Altman and \textbf{(B)} scatter plots of BMD continuous scores predicted by the model from CT scans against DXA T-score BMD ground truth. \textbf{(C)} AUROC of BMD model continuous score against DXA T-score, and \textbf{(D)} AUROC of BMD model binary score against DXA binary classification.\\ AUROC = area under receiver-operator curve}
\end{figure}
\FloatBarrier

 \textit{BMD Subgroup Analyses}
\par
\vspace{0.75em}
\par
 Sensitivity and specificity were computed for all subgroups. The model demonstrated slightly higher sensitivity for male patients than for female patients, although this difference did not reach significance (p = 0.123). There was an inverse relationship observed between sensitivity and specificity as age increased. Younger patients demonstrated the lowest sensitivity and highest specificity, likely reflecting the biological process of osteopenia, where prevalence increases with age. No large discrepancies in model sensitivity were noted between scanner manufacturers. However, specificity with all Philips scanners (69.2\%, 95\% CI 57.8-79.2\%), with the Philips B reconstruction kernel in particular (67.6\%, 95\% CI 55.2-78.5\%), and sensitivity with Toshiba “other” kernels (50\%, 95\% CI 1.3-98.7\%) fell below the 70\% cutoff for acceptable sensitivity/specificity. Specificity for scans with slice thickness $>$3.0 mm (66.7\%, 95\% CI 49.0-81.4\%) was also lower than the cutoff; series with slice thickness $>$3.0 mm are generally uncommon in most current CT imaging protocols. None of the sites exhibited unacceptably low sensitivity or specificity across their subsets (Table 4).

\begingroup
\footnotesize
\setlength{\tabcolsep}{3pt}
\renewcommand{\arraystretch}{0.95}

\setlength{\LTleft}{0pt}
\setlength{\LTright}{0pt}
\setlength{\LTpost}{0pt}

\begin{longtable}{
>{\raggedright\arraybackslash}p{0.36\linewidth}  % Subgroup / label
>{\raggedright\arraybackslash}p{0.14\linewidth}  % Count
>{\raggedright\arraybackslash}p{0.24\linewidth}  % Sensitivity
>{\raggedright\arraybackslash}p{0.21\linewidth}  % Specificity
}

\toprule
\textbf{Subgroup Type} &
\textbf{Count} &
\textbf{Sensitivity (95\% CI)} &
\textbf{Specificity (95\% CI)} \\
\midrule
\endfirsthead

\toprule
\textbf{Subgroup Type} &
\textbf{Count} &
\textbf{Sensitivity (95\% CI)} &
\textbf{Specificity (95\% CI)} \\
\midrule
\endhead

\midrule
\multicolumn{4}{r}{\emph{Continued on next page}}\\
\endfoot

\bottomrule
\addlinespace[6pt]
\caption{Testing data distribution across relevant variables with associated subgroup sensitivity and specificity for the BMD module. All sensitivity and specificity measures are percentages.\\
\emph{* Value fell below the pre-determined 70\% acceptable performance threshold.}}
\label{tab:bmd-performance}
\endlastfoot

Overall & 371 & 81.0\% (74.0--86.8\%) & 78.4\% (72.3--83.7\%) \\
\midrule

\multicolumn{4}{c}{\textbf{Subgroup Analysis by Sex}} \\
\cmidrule(lr){1-4}
Male   & 63  & 94.4\% (72.7--99.9\%) & 75.6\% (60.5--87.1\%) \\
Female & 308 & 79.3\% (71.6--85.7\%) & 79.2\% (72.2--85.0\%) \\
\midrule

\multicolumn{4}{c}{\textbf{Subgroup Analysis by Age}} \\
\cmidrule(lr){1-4}
18--21 & 1   & 100\% (N/A) & N/A \\
22--29 & 5   & 50.0\% (9.5--90.5\%)* & 100\% (43.9--100\%) \\
30--39 & 15  & 50.0\% (18.8--81.2\%)* & 100\% (70.1--100\%) \\
40--49 & 29  & 66.7\% (35.4--87.9\%)* & 95.0\% (76.4--99.1\%) \\
50--69 & 211 & 79.5\% (70.0--87.7\%) & 81.3\% (73.5--87.2\%) \\
70+    & 110 & 90.4\% (79.4--95.8\%) & 62.1\% (49.2--73.4\%)* \\
\midrule

\multicolumn{4}{c}{\textbf{Subgroup Analysis by CT Manufacturer}} \\
\cmidrule(lr){1-4}
GE       & 122 & 77.8\% (62.9--88.8\%) & 79.2\% (68.5--87.6\%) \\
Philips & 151 & 82.2\% (57.8--90.2\%) & 69.2\% (57.8--79.2\%)* \\
Siemens & 66  & 82.4\% (65.5--93.2\%) & 87.5\% (71.0--96.5\%) \\
Toshiba & 32  & 83.3\% (35.9--99.6\%) & 92.3\% (74.9--99.1\%) \\
\midrule

\multicolumn{4}{c}{\textbf{Subgroup Analysis by Kernel}} \\
\cmidrule(lr){1-4}
GE -- STANDARD       & 122 & 77.8\% (62.9--88.8\%) & 79.2\% (68.5--87.6\%) \\
Philips -- B         & 131 & 79.4\% (67.3--88.5\%) & 67.6\% (55.2--78.5\%)* \\
Philips -- Other     & 20  & 100\% (69.2--100\%)   & 80.0\% (44.4--97.5\%) \\
Siemens -- B30f      & 21  & 70.0\% (34.8--93.3\%) & 81.8\% (48.2--97.7\%) \\
Siemens -- Other     & 45  & 87.5\% (67.6--97.3\%) & 90.5\% (69.6--98.8\%) \\
Toshiba -- FC08      & 19  & 100\% (39.8--100\%)   & 93.3\% (68.1--99.8\%) \\
Toshiba -- Other     & 13  & 50.0\% (1.3--98.7\%)* & 90.9\% (58.7--99.8\%) \\
\midrule

\multicolumn{4}{c}{\textbf{Subgroup Analysis by Slice Thickness}} \\
\cmidrule(lr){1-4}
$<2.0$ mm    & 77  & 89.5\% (66.9--98.7\%) & 84.5\% (72.6--92.7\%) \\
2.0--2.5 mm  & 68  & 69.7\% (51.3--84.4\%)* & 82.9\% (66.4--93.4\%) \\
3.0 mm       & 169 & 81.2\% (71.2--88.8\%) & 77.4\% (67.0--85.8\%) \\
$>3.0$ mm    & 57  & 90.5\% (69.6--98.8\%) & 66.7\% (49.0--81.4\%)* \\
\midrule

\multicolumn{4}{c}{\textbf{Subgroup Analysis Across Sites}} \\
\cmidrule(lr){1-4}
MedStar Health                      & 45  & 100\% (78.2--100\%) & 83.3\% (65.3--94.4\%) \\
Diagnósticos da América S.A.        & 94  & 92.0\% (74.0--99.0\%) & 84.1\% (73.3--91.8\%) \\
University Hospitals Cleveland      & 139 & 76.8\% (65.1--86.1\%) & 72.9\% (60.9--82.8\%) \\
University of Alabama at Birmingham & 93  & 75.5\% (61.1--86.7\%) & 75.0\% (59.7--86.8\%) \\

\end{longtable}
\endgroup
\section{\textbf{Discussion}}

In this manuscript, we presented Comp2Comp, a software framework for opportunistic body composition analysis on abdominal CT scans. We transparently describe the clinical endpoints and validation performed to secure FDA 510(k) clearance for the analysis of abdominal aortic aneurysms and bone mineral density within established limits for acceptable clinical use. The AAQ pipeline demonstrated a MAE of 1.58 mm for measuring aortic diameter while the BMD pipeline demonstrated sensitivity of 81.0\% and specificity of 78.4\% for binary classification of low vs. normal bone mineral density.  In both cases, each test dataset was designed to be diverse across institutions, patient characteristics, CT scanners, and disease prevalence. Both described algorithms can enhance patient care without requiring additional radiologist review, exposure to radiation, or costs of new imaging. We open source all code and trained models to improve transparency and reproducibility of our work.
\par
\vspace{1em}
\par
\textit{Clinical Relevance}
\par
\vspace{0.75em}
\par
Approximately 35 million people globally live with an AAA,\textsuperscript{34} and ruptures account for 150,000 deaths annually.\textsuperscript{35,36} Since AAAs usually remain asymptomatic until nearing rupture, early identification of small aneurysms has been a focus for the field of vascular surgery. However, U.S. Preventive Services Task Force (USPSTF) guidelines do not recommend AAA screening for males $<$65 years or women, even though 42\% of ruptures occur in these cohorts,\textsuperscript{37} and even when small AAA is incidentally found on a CT scan, it may not be appropriately flagged and referred. Multiple studies have shown that incidental AAAs remain underreported on abdominopelvic CT scan reports.\textsuperscript{38,39} Opportunistic identification of AAA in unscreened populations may reduce high-mortality AAA ruptures by enabling early lifestyle and therapeutic interventions. Similar approaches for opportunistically assessing coronary artery calcium on chest CT scans have seen recent success.\textsuperscript{9,40} 
\par
\vspace{0.75em}
\par
Beyond augmenting clinical practice, the Comp2Comp AAQ pipeline can help facilitate creating large clinical registries to measure size and progression of aortic measurements. While aortic size may be reported in radiology reports, many reports do not include them\textsuperscript{38} and those that do exhibit high variability in the resulting accuracy.\textsuperscript{41,42} A single, consistent, automated reader that can be applied for all CT scans can help standardize measurements across patients and institutions to create and share AAA registry data. The open-source nature of Comp2Comp can facilitate such clinical research and even analyze imaging in national databases without licensing restrictions.\textsuperscript{43,44} 
\par
\vspace{0.75em}
\par
Osteoporosis affects over 10 million patients in the US alone.\textsuperscript{45} Unlike AAA, it is not in itself a lethal condition; however, the sequela of osteoporosis can lead to severe consequences especially in older patients, including falls, fractures, chronic pain, disability, diminished quality of life, and increased mortality.\textsuperscript{26} Osteoporosis is also a direct driver of increased healthcare costs and person costs due to loss of productivity and informal caregiving.\textsuperscript{46} 
\par
\vspace{0.75em}
\par
The gold standard screening method for osteoporosis using 2D DXA is severely underutilized.\textsuperscript{27,28} Although USPSTF recommends screening all women $>$ 65 years with DXA scan,\textsuperscript{47} only 20.2\% of eligible Medicare beneficiaries undergo DXA scan.\textsuperscript{48} Of patients who sustain a fragility fracture, only 6\% of men have had prior screening,\textsuperscript{49} and only 20\% of women have had a prior osteoporosis diagnosis.\textsuperscript{50}
\par
\vspace{0.75em}
\par
Quantitative measurements of vertebral trabecular bone HU on CT imaging have been validated for osteoporosis screening, but this method has been hindered by the need for radiologists to perform cumbersome manual segmentation of the vertebral bodies.\textsuperscript{25,51,52} Comp2Comp automates the estimation of this validated bone mineral density measure from CT scans without specialized equipment or post-processing to flag patients with osteopenia who are at risk of future osteoporosis. Once identified, patients can be referred for a confirmatory DXA scan and, if necessary, receive appropriate interventions such as pharmacotherapy, lifestyle modifications, and fall prevention strategies. 
\par
\vspace{1em}
\par
\textit{Development process and comparison to prior technologies}
\par
\vspace{0.75em}
\par
The AAQ pipeline represents an improvement on available technologies, two of which have FDA clearance. One such solution, offered by Viz.AI,\textsuperscript{53,54} predicts a binary presence of aneurysm but does not provide actual aortic diameter measurements. Another solution, AIDOC’s BriefCase Aortic Quantification, does provide diameter measurements and is the predicate device for the AAQ 510(k).\textsuperscript{55} However, neither solution is approved for use with non-contrast CT scans. The AAQ algorithm represents the only FDA 510(k)-cleared deep learning solution that provides highly accurate abdominal aortic aneurysm measurements across both contrast and non-contrast CT scans. 
\par
\vspace{0.75em}
\par
The BMD pipeline performance is similar to a prior commercial offering, the Automated Bone Mineral Density (ABMD) Software, developed by HeartLung Corporation, which provides average trabecular bone HU measures.\textsuperscript{56} An estimated Z-score and T-score are then returned to the user, with Pearson r = 0.72 for ABMD performance against DXA T-scores in validation studies. Although Comp2Comp BMD does not return an estimated Z-score or T-score to the user, Pearson r for Comp2Comp BMD continuous CT density score was higher, at r = 0.79. The literature demonstrates that Pearson correlation coefficients for comparing manual CT vertebral bone mineral density estimations against spine DXA T-scores range from r=0.48\textsuperscript{57} to r=0.60,\textsuperscript{58} which places both Comp2Comp BMD and HeartLung ABMD above the range of equivalent manual methods. Because the goal of the Comp2Comp BMD pipeline is to flag high-risk patients and refer them for a future DXA scan, we opted to output a binary flag rather than the continuous score that is less clinically interpretable in the context of broader patient health. 
\par
\vspace{0.75em}
\par
We invite potential users to freely download and deploy the Comp2Comp AAQ and BMD pipelines in their own patient populations to see whether accuracy remains consistent with real world use. Both tools could be used in clinical care and in research studies. Since the code is fully accessible and modifiable, our models and pipelines can also be modularly reorganized and fine-tuned to suit different deployments. We believe that transparently describing our protocols, including pre-determined efficacy endpoints and detailed subgroup analysis, can help end-users develop protocols and even aim for future FDA-clearance of their own algorithms.
\par
\vspace{1em}
\par
\textit{Limitations}
\par
\vspace{0.75em}
\par
Although the AAQ pipeline had high overall performance, its accuracy on patients with aortic endografts was too low (MAE 3.96 mm) for their ultimate inclusion in intended use for the current FDA clearance. For AI-enhanced aortic measurement software, presence of significant luminal thrombus and/or endograft typically requires custom training data aggregation and analysis, beyond the current existing open-source datasets due to the difficulty of this task.\textsuperscript{30,59-64} Although our algorithm works well on a population containing patients with significant endoluminal thrombus, future work can expand the algorithm to postoperative patients and use our pipeline for surveillance across the full spectrum of aortic disease. 
\par
\vspace{0.75em}
\par
The BMD pipeline did not include analysis of contrast-enhanced CT scans, which are common clinically. Future work can expand this algorithm to have similar BMD estimates for arterial, portal-venous, or delayed CT scan phases. While the specificity of the BMD pipeline exceeds the pre-defined threshold (70\%), this threshold may generate false positives when applied to large populations–a common issue for many opportunistic imaging applications.\textsuperscript{65} Further improvement of BMD pipeline specificity would mitigate the number of false positive results. Beyond AAQ and BMD, Comp2Comp also supports body composition analysis for skeletal muscle and visceral and subcutaneous adipose tissue;\textsuperscript{66,67} quantification of aortic calcifications; morphology of solid organs such as the liver, spleen, and kidneys; and automatic identification of contrast phase.\textsuperscript{68} However, these algorithms have not yet been FDA 510(k) cleared. While multi-center validation has been performed for some of these use cases, future work can similarly perform more rigorous analysis. 
\par
\vspace{0.75em}
\par
\textit{Conclusion}
\par
\vspace{0.75em}
\par
The Comp2Comp pipeline includes the first FDA 510(k)-cleared, open-source deep learning solutions for abdominal aortic diameter measurement and vertebral body bone density estimation from abdominopelvic CT scans. By making these technologies fully open-source and publicizing our development and testing process, we enable widespread adoption while fostering continuous community-driven improvements and innovations. Furthermore, by transparently describing all the data submitted to the FDA and the endpoints agreed upon for clearance, we hope to accelerate future research that can be validated by the FDA. While both pipelines have room for improvement, both exceed the necessary accuracy thresholds for current clinical deployment in a manner determined by the FDA. We hope that the successful development of these highly accurate, free-to-use models will set a precedent for further open-source and transparent collaboration in AI for medical imaging.

\section*{Methods}

\textbf{Ethics and data privacy}
\par
\vspace{0.75em}
\par
This study was performed in compliance with relevant ethical standards and local institutional standards for retrospective data processing using AI. All data was handled in accordance with HIPAA safeguards, with IRB waiver of informed consent for fully deidentified data granted through Advarra (IORG 0000635).
\par
\vspace{1em}
\par
\textbf{Predicate Devices:}
\par
\vspace{0.75em}
\par
In the design, evaluation, and approval for AAQ and BMD, we identified one predicate device for each module that had previously achieved FDA 510(k) approval. For AAQ, this was AIDOC Medical: BriefCase Aortic Quantification (K230534),\textsuperscript{55} and for BMD, this was HeartLung Corporation: Automated Bone Mineral Density (ABMD) Software (K213760).\textsuperscript{56}

\textbf{
\subsection{AAQ Deep Learning Pipeline}
}
\par
\vspace{0.75em}
\par
AAQ is a fully automated 3D deep learning pipeline for computing the maximal abdominal aortic diameter from CT scan axial images as outlined in Figure 1A. The system processes input DICOM studies in a sequential manner to ensure anatomical specificity and robustness across the diverse range of clinical images. As a safety measure and to ensure quality control, each stage is contingent on the success of the previous step. 
\par
\vspace{1em}
\par
\textbf{\textit{Step 1: Series Filter}}
\par
\vspace{0.75em}
\par
The pipeline first verifies that CT studies are acquired in axial orientation using the DICOM image orientation patient tag. Any series acquired in coronal or sagittal planes are automatically excluded. This orientation filtering ensures compatibility with the 3D deep learning models used in subsequent segmentation steps.
\par
\vspace{1em}
\par
\textbf{\textit{Step 2: L1-L5 Segmentation and Axial Cropping Model}}
\par
\vspace{0.75em}
\par
Before feeding axial CT scans into our 3D abdominal aorta segmentation algorithm, we perform automated superior-inferior cropping of the input series to the L1-L5 lumbar vertebrae regions. This ensures that aortic segmentations and maximal diameter measurements are solely performed on the abdominal region of the CT scan, excluding the thoracic aorta. The 3D cropping is performed using the open-source TotalSegmentator\textsuperscript{30} spine segmentation nnU-Net\textsuperscript{29} model within the AxialCropper Comp2Comp pipeline.\textsuperscript{69} During this inference stage, if the L1-L5 vertebrae are not detected, the axial series is rejected and the pipeline is exited. If the segmentation is performed successfully, the cropped CT series is passed on to aortic analysis.
\par
\vspace{1em}
\par
\textbf{\textit{Step 3: 3D Abdominal Aorta Segmentation}}
\par
\vspace{0.75em}
\par
\textit{AAQ Model Training Data Acquisition}
\par
\vspace{0.75em}
\par
We trained our custom AAQ 3D algorithm to provide accurate abdominal aortic segmentations from CT scans. Separate cohorts were used to train the AAQ model and test the subsequent AAQ pipeline constructed from the AAQ model to ensure no overlap between training and testing datasets and to demonstrate robust performance in diverse datasets. 
\par
\vspace{0.75em}
\par
The data acquisition process involved building upon existing open-source datasets to train an algorithm capable of classifying not only normal aortas, but also a wide variety of aneurysmal aortas. Our AAQ training dataset consisted of 1804 CT scans. The CT scans containing non-aneurysmal aortas (with a maximal aortic diameter $<$ 3 cm) were sourced from University Hospital Basel in Switzerland (as part of the TotalSegmentator dataset)\textsuperscript{30} and the Longgang District People’s Hospital in Shenzhen, China (part of the AMOS dataset).\textsuperscript{70} 100 CT scans with abnormal aortas were acquired from Stanford University through a retrospective IRB-approved study. These scans consisted of AAAs $>$ 4 cm in diameter and were manually selected for complex morphology including significant mural thrombus and/or endografts when possible. The AAQ training data breakdown across device manufacturer, scanner kVp, and patient demographics is provided in Table 5.

\begingroup
\small
\setlength{\tabcolsep}{3pt}
\renewcommand{\arraystretch}{0.95}
\captionsetup{justification=raggedright,singlelinecheck=false}

\setlength{\LTleft}{0pt}
\setlength{\LTright}{0pt}
\setlength{\LTpost}{0pt}

\begin{longtable}{
>{\raggedright\arraybackslash}p{0.26\linewidth}  % Variable
>{\raggedright\arraybackslash}p{0.32\linewidth}  % Category
>{\raggedright\arraybackslash}p{0.16\linewidth}  % Count (LEFT)
>{\raggedright\arraybackslash}p{0.13\linewidth}  % Percentage
}

\toprule
\textbf{Patient/scan information} & & \textbf{Count (n=1804)} & \textbf{Percentage} \\
\midrule
\endfirsthead

\toprule
\textbf{Patient/scan information} & & \textbf{Count (n=1804)} & \textbf{Percentage} \\
\midrule
\endhead

\bottomrule
\addlinespace[6pt]
\caption{AAQ model training data distribution across relevant variables.}
\label{tab:aaq-training-demographics}
\endlastfoot

\textbf{Site} & University Hospital Basel & 1204 & 66.7\% \\
              & Shenzhen, China           & 500  & 27.7\% \\
              & Stanford University       & 100  & 5.5\% \\
\midrule

\textbf{Sex} & Male   & 1067 & 59.1\% \\
             & Female & 737  & 40.9\% \\
\midrule

\textbf{Age} & 0--19   & 8   & 0.5\% \\
             & 20--29  & 51  & 3.0\% \\
             & 30--39  & 130 & 7.6\% \\
             & 40--49  & 181 & 10.6\% \\
             & 50--59  & 303 & 17.8\% \\
             & 60--69  & 443 & 26.0\% \\
             & 70--79  & 326 & 19.1\% \\
             & 80--89  & 219 & 12.9\% \\
             & 90--100 & 43  & 2.5\% \\
\midrule

\textbf{CT Manufacturer} & Toshiba & 136  & 8.0\% \\
                         & Siemens & 1128 & 66.2\% \\
                         & GE      & 272  & 16.0\% \\
                         & Philips & 169  & 9.8\% \\
\midrule

\textbf{Kilovoltage Peak (kVp)} & 80  & 20  & 1.7\% \\
                               & 100 & 400 & 33.2\% \\
                               & 120 & 600 & 49.8\% \\
                               & 140 & 184 & 15.3\% \\

\end{longtable}
\endgroup

\textit{AAQ Model Training and Integration}
\par
\vspace{0.75em}
\par
Within the AAQ pipeline, the L1-L5 cropped volume is passed into the AAQ nnU-Net model to begin the aortic analysis stage. Model training was performed in accordance with the TotalSegmentator setup for 4,000 epochs. Data augmentations followed the full-resolution nnU-Net protocol, with all forms of mirroring explicitly disabled to preserve the anatomical orientation of the aorta. The nnU-Net framework automatically configured the optimal patch size, network topology, and training schedule based on dataset properties which enabled efficient learning without manual tuning.\textsuperscript{29} To combat the dataset bias towards normal aortas and increase the representation of abnormal aortas, we performed raw data-level upsampling of the Stanford cohort by triplicating each sample in the training set. The model was therefore trained on a dataset in which Stanford cohort samples simply appeared three times more frequently. The usage of this cohort significantly improved segmentation accuracy and measurement performance as shown in Supplementary Figure 1, where the model developed the ability to accurately identify the aortic wall instead of the contrast lumen for the same test scan.
\par
\vspace{1em}
\par
\textbf{\textit{Step 4: Maximal Axial Diameter Selection}}
\par
\vspace{0.75em}
\par
We then compute the maximal diameter from the 3D abdominal aorta segmentation. The 3D segmentation data output from the AAQ model is converted to individual 2D segmentation masks that are overlayed in the axial plane. Each 2D segmentation mask on each axial slice is fitted with an ellipse. This enables the algorithm to compute and draw the minor axis and subsequently perform slice-by-slice measurements. The minor axis is measured in DICOM pixels and stored for each corresponding slice. After measurements are completed for each 2D axial mask, the largest minor axis diameter, representing the maximal aortic diameter, is output along with the corresponding slice number to be used in the subsequent step. The largest minor axis is chosen to measure aortic diameter, as opposed to largest major axis, to mitigate elliptical “stretching” of the diameter along the major axis when the aorta travels oblique to the axial cuts (Figure 5).

\begin{figure}[htbp]
  \centering
  \includegraphics[width=\linewidth]{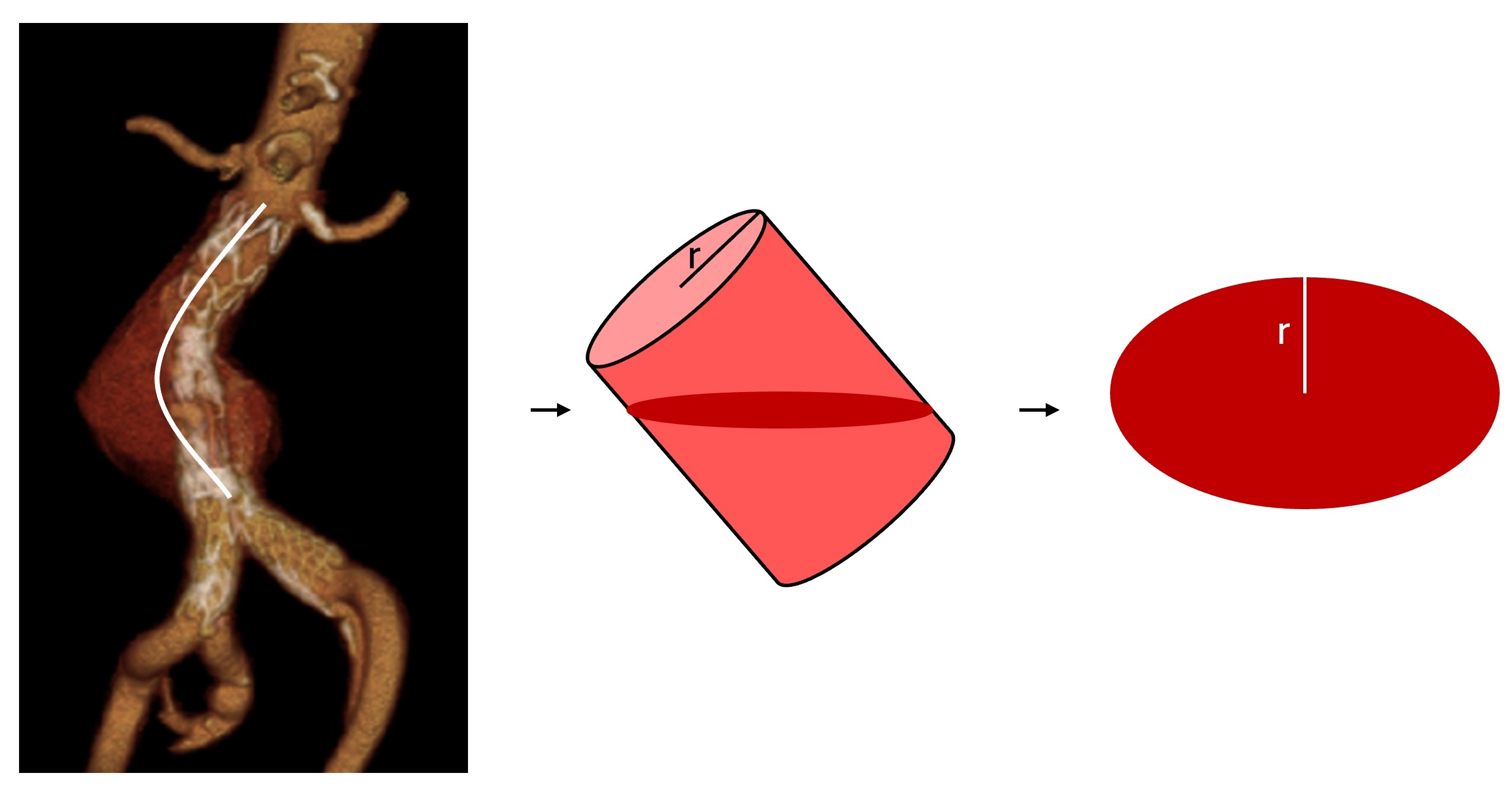}
  \caption{Selection of largest minor axis of cross-sectional aortic ellipse to accurately represent true aortic diameter.}
\end{figure}
\FloatBarrier

When the aorta courses oblique to the axial slices of a CT scan, axial cross sections of a cylindrical aorta will appear as ellipses, not circles, elongated along the direction of aortic traversal. The minor axis of the ellipse remains equal to the radius of the aorta modeled as a cylinder, no matter the angulation of the centerline relative to the axial slices.
\par
\vspace{0.75em}
\par
It must be noted that this assumption only holds true for aneurysms that can be roughly modeled as a cylinder. Saccular aneurysms (where the aortic wall expands on only one side of the vessel) cannot be modeled as a cylinder, and use of the largest minor axis will not give an accurate estimate of centerline-orthogonal largest diameter. We therefore excluded saccular aneurysms from the training set, testing set, and indications for use (IFU) of the pipeline. 
\par
\vspace{1em}
\par
\textbf{\textit{Step 5: Voxel to mm Conversion}}
\par
\vspace{0.75em}
\par
Voxel measurements are converted to clinician interpretable millimeters and centimeters using voxel header metadata sourced from the original DICOM series (\textit{DICOM }\textit{Input CT Series} in Figure 1). The voxel spacing header is consistently found in DICOM acquisition and provides a factor which can be multiplied by the raw voxel measurement to derive a human-readable measurement.
\par
\vspace{1em}
\par
\textbf{\textit{Step 6: Explainable Output Generation}}
\par
\vspace{0.75em}
\par
The core output is the maximal abdominal aortic diameter and the slice number of the widest portion of the aorta. To allow for clinical interpretation, the AAQ pipeline has multiple additional outputs which were not evaluated as part of the FDA 510(k) clearance. The pipeline generates a segmentation visualization of the maximum diameter AAA slice, a histogram of aortic diameter progression through the axial slices, as well as a video showing the aortic segmentation progression. These outputs are configurable and can be used to enable troubleshooting and manual verification of measurements if necessary.
\par
\vspace{1em}
\par
\textbf{AAQ Evaluation Data Acquisition}
\par
\vspace{0.75em}
\par
Our AAQ deep learning pipeline was evaluated on external multi-institutional data. Three cohorts were geographically distributed throughout the United States and distinct from the cohort used during training: MedStar Health, Maryland; North Carolina (acquired from data broker); University of Alabama at Birmingham, Alabama. One cohort was international: Diagnósticos da América S.A. (DASA), Brazil. CT imaging patients were selected for diversity of sex, age, and aortic characteristics, taking care to ensure that $\sim$50\% of the cohort was female and at least 30\% of the scans contained an abdominal aortic aneurysm. For each unique patient, either a single noncontrast CT or IV contrast-enhanced CT was selected to ensure balanced proportions of slice size, contrast use, patients, and CT manufacturers across sites. Each site would represent at least 10\% of the total data, but no single site would represent greater than 50\% of the total data. We excluded CT scans that were obtained postoperatively from endovascular aortic surgery from the primary analysis and investigated these separately after finding that model performance in postoperative scans was not yet satisfactory for clinical use.
\par
\vspace{0.75em}
\par
A pilot study was conducted on 32 patients to project necessary sample size based on preliminary performance of AAQ. This pilot study followed the same clinical study design as this pivotal clinical study and had a MAE of 1.58 mm (SD 1.09 mm). Power analysis was conducted to ensure the upper bound of a 95\% CI for the MAE of the pivotal trial did not exceed 2.0 mm at 80\% power and two-sided alpha = 0.05. Consequently, the sample size to meet the acceptance criteria was calculated to be at least 29 patients. A sample size of at least 250 patients was targeted to be conservative and to ensure stratification across important sub-groups.
\par
\vspace{1em}
\par
\textbf{Pipeline Evaluation Study}
\par
\vspace{0.75em}
\par
\textbf{\textit{Ground-Truth Determination}}
\par
\vspace{0.75em}
\par
AAQ measurements of largest abdominal aortic diameter were compared against ground truth measurements from the same scans, measured by three US Board Certified Radiologists. The three radiologists measured the largest minor axis diameter of each scan in parallel to each other in a blinded manner. Each measurement was then reviewed for compliance against a standardized set of measurement instructions by a Bunkerhill Health employee. Each radiologist would then review any compliance commentary (if present) and choose to adjust their measurement or reject the comment. After this process was repeated for each radiologist across each scan, the average of the three measurements was calculated and established as the ground truth in this study.
\par
\vspace{1em}
\par
\textbf{\textit{Statistical Analysis Metrics}}
\par
\vspace{0.75em}
\par
To compare the statistical concordance between the board-certified radiologists (ground truth) and the AAQ pipeline, we use two metrics: mean absolute error (MAE) and intraclass correlation coefficient (ICC). Mean absolute error was calculated as the average of absolute values of difference between each CT’s AAQ output and the radiologist ground truth measurement. We chose 2.0 mm as the acceptability threshold for MAE based both on FDA clearance criteria for predicate devices, and because 2.0 mm is considered a clinically reasonable threshold for interobserver variability of manual AAA measurements.\textsuperscript{32} This same study noted that 35\% of interobserver pairs of AAA measurements differed by $>$ 2mm, and 17\% differed by $>$ 5 mm.\textsuperscript{32} For the Bland-Altman analysis, 5 mm was therefore chosen as the threshold of acceptability for visual concordance.
\par
\vspace{0.75em}
\par
ICC with 95\% confidence interval was calculated for radiologist measurements alone, and then again for radiologist measurements combined with AAQ measurement, to determine how closely the measurements for each scan clustered with and without AAQ output. The ICC enabled us to understand the impact of AAQ in improving the consistency and reliability of measurements. The acceptance criterion for this endpoint was set at the upper limit for the confidence intervals of the difference in the two ICCs being less than 0.05.
\par
\vspace{0.75em}
\par
\textbf{Data and Code Availability}
\par
\vspace{0.75em}
\par
The data used in this study was acquired in an agreement with the involved hospitals for usage in training of the AAQ model and ultimately the full AAQ pipeline. Therefore, the data is not made publicly available. The AAQ pipeline, AAQ model, and all associated code are publicly available through the Comp2Comp\textsuperscript{69} pipeline on GitHub, with the model files hosted on the HuggingFace platform. Through integration within the Comp2Comp ecosystem, AAQ is available for versatile usage and can be run on batches of DICOM studies to create large-scale registries for research studies. We also integrated the pipeline within a front-end web application hosted on HuggingFace spaces which allows for DICOM anonymization, upload, and output visualization in the web browser with an NVIDIA T4 medium partition GPU (8 vCPU, 30 GB RAM, 16 GB VRAM).
\textbf{
\subsection{BMD Deep Learning Pipeline}
}
\par
\vspace{0.75em}
\par
The BMD pipeline consists of seven steps, as illustrated in Figure 2A and detailed below.
\par
\vspace{0.75em}
\par
\textbf{Step 1:} The input DICOM file undergoes quality control to verify the following requirements: The image is Original (not processed or edited), Primary (not created from another image), Axial (slices are reconstructed in the axial direction), and it is acquired at 120 kVp. The following is expected to hold for the CT scan but is not explicitly checked by the algorithm: The CT reconstruction kernel should be a soft type (see Appendix 2), slice thickness should be less than or equal to 5 mm, and the CT should not be a pediatric scan.
\par
\vspace{0.75em}
\par
\textbf{Step 2:} A spine segmentation model identifies and segments vertebrae L1–L4. Scans in which one or more of these vertebrae cannot be detected are excluded from analysis.
\par
\vspace{0.75em}
\par
\textbf{Step 3:} For each vertebral level (L1–L4), a three-dimensional region of interest (ROI) is defined to sample trabecular bone. This ROI is then used to extract the mean vertebra HU from the CT scan.
\par
\vspace{0.75em}
\par
\textbf{Step 4:} Visceral adipose tissue (VAT) is segmented, and an air calibration ROI is positioned 20 mm anterior to the most anterior VAT voxel. The air ROI dimensions are 20 mm (anterior-posterior) × 50 mm (lateral) × full scan length (superior-inferior).
\par
\vspace{0.75em}
\par
\textbf{Step 5:} Quality control is performed by evaluating the mean Hounsfield unit (HU) value within the air ROI. Scans are excluded if this value falls below -1050 HU or exceeds -950 HU. For scans passing quality control, a two-point calibration is established using VAT (assumed to be -95 HU) and air (assumed to be -1000 HU) to derive slope and intercept parameters for HU recalibration.
\par
\vspace{0.75em}
\par
\textbf{Step 6:} The HU values extracted from the central ROIs of vertebrae L1–L4 are recalibrated using the slope and intercept calculated in Step 5.
\par
\vspace{0.75em}
\par
\textbf{Step 7:} The mean recalibrated HU value across the L1–L4 vertebral ROIs is used to predict osteoporosis risk. A mean HU value below 300 indicates a predicted T-score less than -1 (osteopenic or osteoporotic), while values at or above 300 indicate a predicted T-score of -1 or greater (normal). The threshold of 300 HU was chosen using a cohort from UCSF Medical Center (separate from evaluation cohorts). The pipeline also generates supplementary outputs including a DXA-equivalent score and estimated T- and Z-scores derived from the mean recalibrated HU values. Conversion from Z-score to T-scores was done with the conversion factor from the NHANES study.\textsuperscript{71} However, only the binary classification (T-score above or below -1) has received FDA clearance; the supplementary metrics are not FDA cleared.
\par
\vspace{1em}
\par
\textbf{BMD Evaluation Data Acquisition}
\par
\vspace{0.75em}
\par
The BMD deep learning pipeline was evaluated on external multi-institutional data collected from four geographically diverse sites: MedStar Health, Maryland; University Hospitals Cleveland Medical Center, Ohio; University of Alabama at Birmingham, Alabama, and Diagnósticos da América S.A. (DASA), Brazil. Each participating institution was given an identical request to identify pairs of abdominal CT scans and DXA scans in adult patients, obtained between 2014 and 2024. Abdominal CT scans had to contain a non-contrast series, and DXA scans had to have an associated report in which L1-L4 bone density, T-score, and Z-score were present. The time difference between CT and DXA scan could not exceed 6 months, and in the event of multiple scans, the closest interval pair was to be selected.
\par
\vspace{0.75em}
\par
Once study pairs were identified by collaborating institutions, a secondary quality control step was completed, where we verified that each DXA report had measurements of L1-L4 vertebral body bone density and that each CT scan contained a non-contrast series. A third quality control step involved checking if each included CT scan was performed at 120 kVp with standard or soft reconstruction kernels, with slice thickness $\le$ 5.0 mm in the axial orientation. At each stage, patients not meeting requirements were eliminated (Supplemental Figure 2).
\par
\vspace{0.75em}
\par
A pilot study was conducted to evaluate preliminary performance of Bunkerhill BMD and provide a power calculation for the pivotal trial. This pilot study followed the same clinical study design as this pivotal clinical and was conducted on 52 unique studies from Emory University Medical Center. The power analysis was conducted to achieve a target sensitivity and specificity of 0.70 at an 80\% power and alpha = 0.05. Consequently, the sample size to meet the acceptance criteria was calculated to be at least 158 studies; to be conservative, 350 was set as a target enrollment level to ensure stratification across important subgroups. No specific parameter was set for the proportion of patients with low bone density. Ultimately, the test dataset had a ground truth prevalence of 42.6\% low bone density, which adheres closely to the prevalence of low bone density in US adults $>$50 years (43.1\%).\textsuperscript{72} 
\par
\vspace{1em}
\par
\textbf{Pipeline Evaluation Study}
\par
\vspace{0.75em}
\par
\textbf{\textit{Ground-Truth Determination}}
\par
\vspace{0.75em}
\par
Each patient was considered to have a ground-truth of low bone density if the composite DXA scan T-score estimated from the L1-L4 bone density found in the DXA scan report was $\le$-1.0. T-score $\le$-1.0 corresponds with osteopenia, and so we chose this threshold to catch not only patients with frank osteoporosis, but also those at risk of developing osteoporosis without intervention.
\par
\vspace{0.75em}
\par
\textbf{\textit{Statistical Analysis Metrics}}
\par
\vspace{0.75em}
\par
To compare the statistical concordance between DXA T-score and the BMD pipeline, we evaluated sensitivity and specificity of the BMD binary classification against the ground truth classification generated from DXA T-score threshold of less than -1.0, corresponding to osteopenia. We reviewed the literature and found that for most evaluations of manual estimation of bone density from CT scan, sensitivity and specificity were $\sim$70\% when compared against DXA scan ground truth.\textsuperscript{51,52,73,74} We therefore set 70\% as the acceptability threshold for sensitivity and specificity. 
\par
\vspace{0.75em}
\par
We also decided to evaluate the model’s intermediate output of continuous bone mineral density estimation, since prior technologies\textsuperscript{56} offer this output to users. Since HeartLung ABCT found a Pearson correlation coefficient (PCC) of r = 0.72 when comparing their software’s continuous scores with DXA T-score ground truth scores, we set our acceptance criterion for PCC at r $>$ 0.70. 
\par
\vspace{1em}
\par
\textbf{Data and Code Availability}
\par
\vspace{0.75em}
\par
The data used in this study was acquired in an agreement with the involved hospitals for usage in training of the BMD pipeline. Therefore, the data is not made publicly available. The BMD pipeline and all associated code is available through the Comp2Comp\textsuperscript{69} pipeline on GitHub. 

\clearpage

\textbf{
\section{Supplemental Figures:}
}
\setcounter{table}{0}
\setcounter{figure}{0}

\renewcommand{\thetable}{S\arabic{table}}
\renewcommand{\thefigure}{S\arabic{figure}}

\begin{table}[ht]
\small
\setlength{\tabcolsep}{4pt}
\renewcommand{\arraystretch}{0.95}
\centering

\begin{tabular}{
>{\raggedright\arraybackslash}p{0.10\linewidth}  % Site
>{\raggedright\arraybackslash}p{0.13\linewidth}  % White
>{\raggedright\arraybackslash}p{0.13\linewidth}  % Black
>{\raggedright\arraybackslash}p{0.13\linewidth}  % Asian
>{\raggedright\arraybackslash}p{0.20\linewidth}  % Hispanic
>{\raggedright\arraybackslash}p{0.20\linewidth}  % Other
}

\toprule
\textbf{Site} &
\textbf{\% White} &
\textbf{\% Black} &
\textbf{\% Asian} &
\textbf{\% Hispanic/Latino} &
\textbf{\% Other/Unknown} \\
\midrule

MedStar & 41\% & 24\% & 11\% & 18\% & 6\% \\
DASA    & 54.3\% & 10.1\% & N/A & N/A & 35.6\% \\
UH      & 36.7\% & 46.8\% & 2.3\% & 12.8\% & 1.4\% \\
UAB     & 25.6\% & 67.1\% & 1.4\% & 4.9\% & 3.6\% \\

\bottomrule
\end{tabular}

\caption{Race/ethnicity of overall patient populations at each site participating in the BMD validation study.}
\label{tab:race-ethnicity-sites}
\end{table}

\begin{figure}[htbp]
  \centering
  \includegraphics[width=\linewidth]{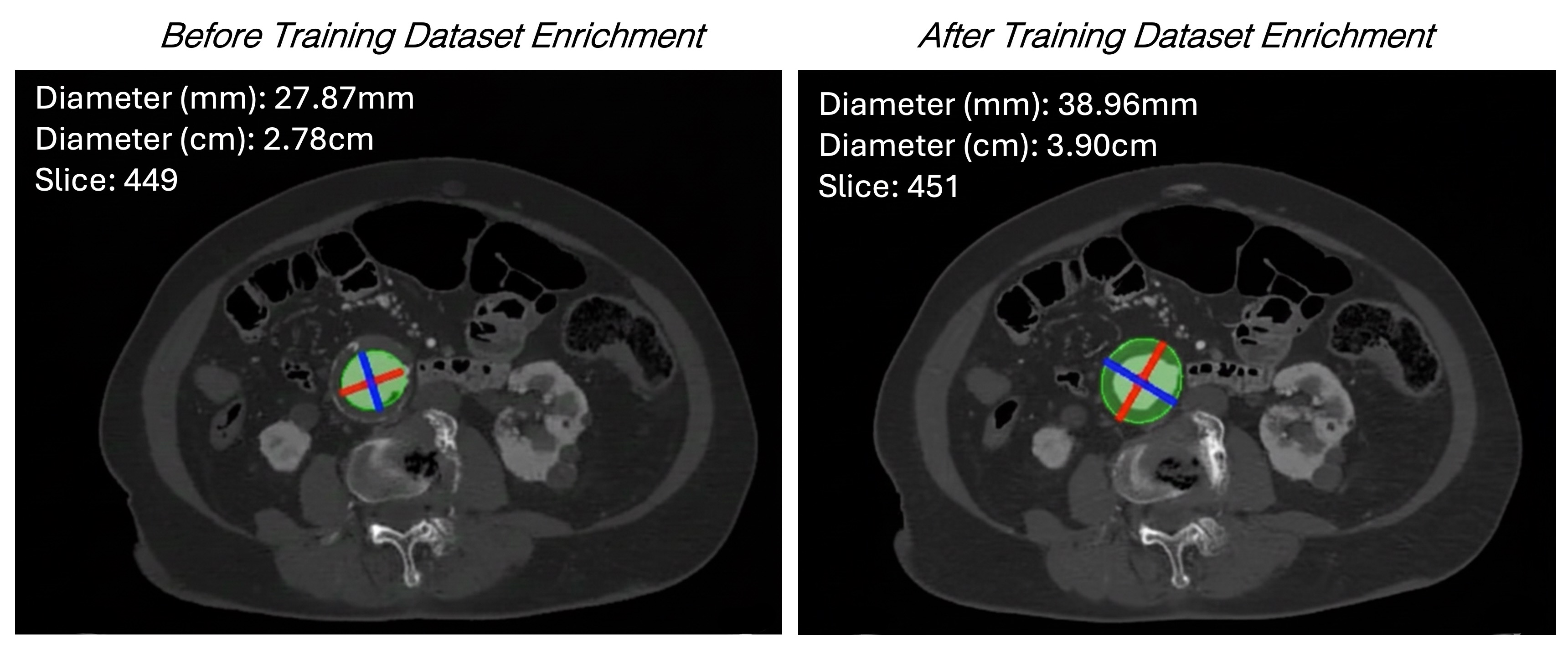}
  \caption{Aortic segmentation and diameter extraction on an aorta with thrombus after only training on open-source datasets (left) and after enriching training with Stanford aortic aneurysm cohort upsampling (right).}
\end{figure}
\FloatBarrier

\FloatBarrier

\noindent
\begin{minipage}{\textwidth}
  \centering
  \includegraphics[width=\linewidth]{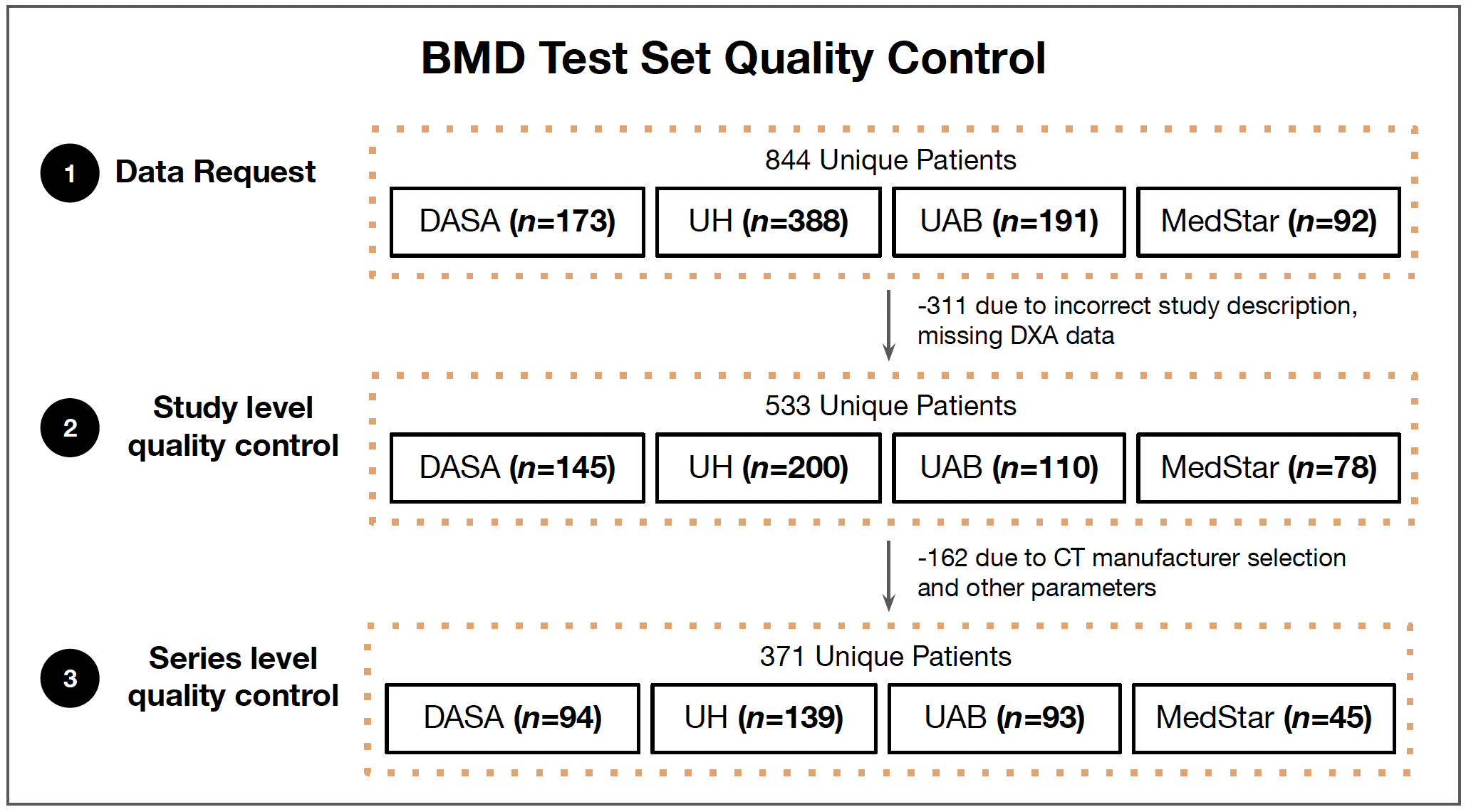}
  \captionsetup{justification=raggedright,singlelinecheck=false}
  \captionof{figure}{Quality control flow chart for patient inclusion in BMD test set.}
  \label{fig:bmd-qc}
\end{minipage}
\FloatBarrier

\clearpage

\section{\textbf{References}}

1. Food US, Drug A. Artificial Intelligence-Enabled Medical Devices. Accessed January 2026

2. Pickhardt PJ, Correale L, Hassan C. AI-based opportunistic CT screening of incidental cardiovascular disease, osteoporosis, and sarcopenia: cost-effectiveness analysis. \textit{Abdom Radiol}. 2023;doi:10.1007/s00261-023-03800-9

3. Mervak BM, Fried JG, Wasnik AP. A Review of the Clinical Applications of Artificial Intelligence in Abdominal Imaging. \textit{Diagnostics (Basel)}. Sep 8 2023;13(18) \\doi:10.3390/diagnostics13182889

4. Blankemeier L, Cohen JP, Kumar A, Van Veen D, Gardezi SJS, Paschali M, et al. Merlin: A Vision Language Foundation Model for 3D Computed Tomography. \textit{arXiv}. 2024; \\doi:10.48550/arXiv.2406.06512

5. Smith-Bindman R, Chu PW, Azman Firdaus H, Stewart C, Malekhedayat M, Alber S, et al. Projected Lifetime Cancer Risks From Current Computed Tomography Imaging. \textit{JAMA Internal Medicine}. 2025;185(6):710–719. doi:10.1001/jamainternmed.2025.0505

6. Aali A, Johnston A, Blankemeier L, Van Veen D, Derry LT, Svec D, et al. Automated detection of underdiagnosed medical conditions via opportunistic imaging. \textit{arXivorg}. 2024; \\doi:10.48550/ARXIV.2409.11686

7. Boutin RD, Lenchik L. Value-Added Opportunistic CT: Insights Into Osteoporosis and Sarcopenia. \textit{American Journal of Roentgenology}.  2020;215(3):582–594. doi:10.2214/ajr.20.22874

8. Pickhardt PJ. Value-added Opportunistic CT Screening: State of the Art. \textit{Radiology}. \\2022;303(2):241–254. doi:10.1148/radiol.211561

9. Sandhu AT, Rodriguez F, Ngo S, Patel BN, Mastrodicasa D, Eng D, et al. Incidental Coronary Artery Calcium: Opportunistic Screening of Previous Nongated Chest Computed Tomography Scans to Improve Statin Rates (NOTIFY-1 Project). \textit{Circulation}. Feb 28 2023;147(9):703–714. \\doi:10.1161/circulationaha.122.062746

10. Eltawil FA, Atalla M, Boulos E, Amirabadi A, Tyrrell PN. Analyzing Barriers and Enablers for the Acceptance of Artificial Intelligence Innovations into Radiology Practice: A Scoping Review. \textit{Tomography}. Jul 28 2023;9(4):1443–1455. doi:10.3390/tomography9040115

11. Chatterjee N, Duda J, Gee J, Elahi A, Martin K, Doan V, et al. A Cloud-Based System for Automated AI Image Analysis and Reporting. \textit{J Digit Imaging Inform med}. 2024/7/31/ 2024;38(1):368–379. doi:10.1007/s10278-024-01200-z

12. Aggarwal V, Maslen C, Abel RL, Bhattacharya P, Bromiley PA, Clark EM, et al. Opportunistic diagnosis of osteoporosis, fragile bone strength and vertebral fractures from routine CT scans; a review of approved technology systems and pathways to implementation. \textit{Therapeutic Advances in Musculoskeletal}. 2021;13doi:10.1177/1759720x211024029

13. Greenhalgh T, Wherton J, Papoutsi C, Lynch J, Hughes G, A'Court C, et al. Beyond Adoption: A New Framework for Theorizing and Evaluating Nonadoption, Abandonment, and Challenges to the Scale-Up, Spread, and Sustainability of Health and Care Technologies. \textit{J Med Internet Res}. 2017/11/01 2017;19(11):e367. doi:10.2196/jmir.8775

14. Food US, Drug A. Artificial Intelligence‑Enabled Medical Devices. 2025;

15. Benjamens S, Dhunnoo P, Meskó B. The state of artificial intelligence-based FDA-approved medical devices and algorithms: an online database. \textit{npj Digital Medicine}. 2020/09/11 2020;3(1):118. \\doi:10.1038/s41746-020-00324-0

16. Harish KB, Price WN, Aphinyanaphongs Y. Open-Source Clinical Machine Learning Models: Critical Appraisal of Feasibility, Advantages, and Challenges. \textit{JMIR Form Res}. Apr 11 2022;6(4):e33970. doi:10.2196/33970

17. Marey A, Arjmand P, Alerab ADS, Eslami MJ, Saad AM, Sanchez N, et al. Explainability, transparency and black box challenges of AI in radiology: impact on patient care in cardiovascular radiology. \textit{Egyptian Journal of Radiology and Nuclear Medicine}. 2024/09/13 2024;55(1):183. doi:10.1186/s43055-024-01356-2

18. Muralidharan V, Adewale BA, Huang CJ, Nta MT, Ademiju PO, Pathmarajah P, et al. A scoping review of reporting gaps in FDA-approved AI medical devices. \textit{npj Digital Medicine}. 2024/10/03 2024;7(1):273. doi:10.1038/s41746-024-01270-x

19. Omoumi P, Richiardi J. Independent Evaluation of Commercial Diagnostic AI Solutions: A Necessary Step toward Increased Transparency. \textit{Radiology}. 2024/01/01 2024;310(1):e233299. \\doi:10.1148/radiol.233299

20. Nair M, Svedberg P, Larsson I, Nygren JM. A comprehensive overview of barriers and strategies for AI implementation in healthcare: Mixed-method design. \textit{PLOS ONE}. 2024;19(8):e0305949. \\doi:10.1371/journal.pone.0305949

21. Trivedi H, Khosravi B, Gichoya J, Benson L, Dyckman D, Galt J, et al. AI in Action: A Road Map From the Radiology AI Council for Effective Model Evaluation and Deployment. \textit{Journal of the American College of Radiology}. 2025/09/01 2025;22(9):1041–1049. \\doi:\href{https://doi.org/10.1016/j.jacr.2025.05.016}{https://doi.org/10.1016/j.jacr.2025.05.016}

22. Savage CH, Tanwar M, Elkassem AA, Sturdivant A, Hamki O, Sotoudeh H, et al. Prospective Evaluation of Artificial Intelligence Triage of Intracranial Hemorrhage on Noncontrast Head CT Examinations. \textit{American Journal of Roentgenology}. 2024;223(5):e2431639. \\doi:10.2214/ajr.24.31639

23. Wu E, Wu K, Daneshjou R, Ouyang D, Ho DE, Zou J. How medical AI devices are evaluated: limitations and recommendations from an analysis of FDA approvals. \textit{Nature Medicine}. 2021;doi:10.1038/s41591-021-01312-x

24. Krafcik BM, Stone DH, Cai M, Jarmel IA, Eid M, Goodney PP, et al. Changes in global mortality from aortic aneurysm. \textit{Journal of Vascular Surgery}. 2024/07/01/ 2024;80(1):81–88.e1. \\doi:\href{https://doi.org/10.1016/j.jvs.2024.02.025}{https://doi.org/10.1016/j.jvs.2024.02.025}

25. Brett AD, Brown JK. Quantitative computed tomography and opportunistic bone density screening by dual use of computed tomography scans. \textit{J Orthop Translat}. Oct 2015;3(4):178–184. \\doi:10.1016/j.jot.2015.08.006

26. Bliuc D, Nguyen ND, Nguyen TV, Eisman JA, Center JR. Compound risk of high mortality following osteoporotic fracture and refracture in elderly women and men. \textit{J Bone Miner Res}. Nov 2013;28(11):2317–24. doi:10.1002/jbmr.1968

27. Amarnath AL, Franks P, Robbins JA, Xing G, Fenton JJ. Underuse and Overuse of Osteoporosis Screening in a Regional Health System: a Retrospective Cohort Study. \textit{J Gen Intern Med}. Dec 2015;30(12):1733–40. doi:10.1007/s11606-015-3349-8

28. Alswat K, Adler SM. Gender differences in osteoporosis screening: retrospective analysis. \textit{Arch Osteoporos}. 2012;7:311–3. doi:10.1007/s11657-012-0113-0

29. Isensee F, Jaeger PF, Kohl SAA, Petersen J, Maier-Hein KH. nnU-Net: a self-configuring method for deep learning-based biomedical image segmentation. \textit{Nature Methods}. 2021/02/01 2021;18(2):203–211. doi:10.1038/s41592-020-01008-z

30. Wasserthal J, Breit H-C, Meyer MT, Pradella M, Hinck D, Sauter AW, et al. TotalSegmentator: Robust Segmentation of 104 Anatomic Structures in CT Images. \textit{Radiology: Artificial Intelligence}. 2023/09/01 2023;5(5):e230024. doi:10.1148/ryai.230024

31. Sekuboyina A, Husseini ME, Bayat A, Löffler M, Liebl H, Li H, et al. VerSe: A Vertebrae labelling and segmentation benchmark for multi-detector CT images. \textit{Medical Image Analysis}. 2021/10/01/ 2021;73:102166. doi:\href{https://doi.org/10.1016/j.media.2021.102166}{https://doi.org/10.1016/j.media.2021.102166}

32. Lederle FA, Wilson SE, Johnson GR, Reinke DB, Littooy FN, Acher CW, et al. Variability in measurement of abdominal aortic aneurysms. Abdominal Aortic Aneurysm Detection and Management Veterans Administration Cooperative Study Group. \textit{J Vasc Surg}. Jun 1995;21(6):945–52. doi:10.1016/s0741-5214(95)70222-9

33. Dimai HP. Use of dual-energy X-ray absorptiometry (DXA) for diagnosis and fracture risk assessment; WHO-criteria, T- and Z-score, and reference databases. \textit{Bone}. Nov 2017;104:39–43. doi:10.1016/j.bone.2016.12.016

34. Song P, He Y, Adeloye D, Zhu Y, Ye X, Yi Q, et al. The Global and Regional Prevalence of Abdominal Aortic Aneurysms: A Systematic Review and Modeling Analysis. \textit{Ann Surg}. Jun 1 2023;277(6):912–919. doi:10.1097/sla.0000000000005716

35. Yang W, Wu S, Qi F, Zhao L, Hai B, Dong H, et al. Global trends and stratified analysis of aortic aneurysm mortality: insights from the GBD 2021 study. \textit{Front Cardiovasc Med}. 2025;12:1496166. doi:10.3389/fcvm.2025.1496166

36. Reimerink JJ, van der Laan MJ, Koelemay MJ, Balm R, Legemate DA. Systematic review and meta-analysis of population-based mortality from ruptured abdominal aortic aneurysm. \\\textit{Br J Surg}. Oct 2013;100(11):1405–13. doi:10.1002/bjs.9235

37. Abdulameer H, Al Taii H, Al-Kindi SG, Milner R. Epidemiology of fatal ruptured aortic aneurysms in the United States (1999-2016). \textit{J Vasc Surg}. Feb 2019;69(2):378–384.e2. \\doi:10.1016/j.jvs.2018.03.435

38. Claridge R, Arnold S, Morrison N, van Rij AM. Measuring abdominal aortic diameters in routine abdominal computed tomography scans and implications for abdominal aortic aneurysm screening. \textit{Journal of Vascular Surgery}. 2017/06/01/ 2017;65(6):1637–1642. \\doi:\href{https://doi.org/10.1016/j.jvs.2016.11.044}{https://doi.org/10.1016/j.jvs.2016.11.044}

39. Gordon JRS, Wahls T, Carlos RC, Pipinos II, Rosenthal GE, Cram P. Failure to Recognize Newly Identified Aortic Dilations in a Health Care System With an Advanced Electronic Medical Record. \textit{Annals of Internal Medicine}. 2009/07/07 2009;151(1):21–27. doi:10.7326/0003-4819-151-1-200907070-00005

40. Dudum R, Jain SS, Mastrodicasa D, Furst A, Xu S, Ngo S, et al. Effects of Real-Time Notification of AI-Detected Incidental Coronary Artery Calcium on Statin Prescription: the NOTIFY-PICTURE Trial. \textit{Circulation}. 0(0)doi:10.1161/CIRCULATIONAHA.125.078155

41. Jaakkola P, Hippeläinen M, Farin P, Rytkönen H, Kainulainen S, Partanen K. Interobserver variability in measuring the dimensions of the abdominal aorta: comparison of ultrasound and computed tomography. \textit{Eur J Vasc Endovasc Surg}. Aug 1996;12(2):230–7. doi:10.1016/s1078-5884(96)80112-2

42. Cayne NS, Veith FJ, Lipsitz EC, Ohki T, Mehta M, Gargiulo N, et al. Variability of maximal aortic aneurysm diameter measurements on CT scan: significance and methods to minimize. \textit{Journal of Vascular Surgery}. 2004;39(4):811–815. doi:10.1016/j.jvs.2003.11.042

43. Gaziano JM, Concato J, Brophy M, Fiore L, Pyarajan S, Breeling J, et al. Million Veteran Program: A mega-biobank to study genetic influences on health and disease. \textit{J Clin Epidemiol}. Feb 2016;70:214–23. doi:10.1016/j.jclinepi.2015.09.016

44. Cronenwett JL, Kraiss LW, Cambria RP. The Society for Vascular Surgery Vascular Quality Initiative. \textit{Journal of Vascular Surgery}. 2012/05/01/ 2012;55(5):1529–1537. \\doi:\href{https://doi.org/10.1016/j.jvs.2012.03.016}{https://doi.org/10.1016/j.jvs.2012.03.016}

45. Wright NC, Looker AC, Saag KG, Curtis JR, Delzell ES, Randall S, et al. The recent prevalence of osteoporosis and low bone mass in the United States based on bone mineral density at the femoral neck or lumbar spine. \textit{J Bone Miner Res}. Nov 2014;29(11):2520–6. \\doi:10.1002/jbmr.2269

46. Williams SA, Daigle SG, Weiss R, Wang Y, Arora T, Curtis JR. Economic Burden of Osteoporosis-Related Fractures in the US Medicare Population. \textit{Ann Pharmacother}. Jul 2021;55(7):821–829. doi:10.1177/1060028020970518

47. Compston JE, McClung MR, Leslie WD. Osteoporosis. \textit{The Lancet}. 2019;393(10169):364–376. doi:10.1016/S0140-6736(18)32112-3

48. Gyftopoulos S, Pelzl CE, Chang CY. Quantifying the Opportunity and Economic Value of Bone Density Screening Using Opportunistic CT: A Medicare Database Analysis. \textit{Journal of the American College of Radiology}. 2025;22(3):349–357. doi:10.1016/j.jacr.2024.10.003

49. Williams SA, Daigle SG, Weiss R, Wang Y, Arora T, Curtis JR. Characterization of Older Male Patients with a Fragility Fracture. \textit{Arthritis \& Rheumatology}. 2020/11/07 2020;72(Suppl 10)

50. Weaver J, Sajjan S, Lewiecki EM, Harris ST. Diagnosis and Treatment of Osteoporosis Before and After Fracture: A Side-by-Side Analysis of Commercially Insured and Medicare Advantage Osteoporosis Patients. \textit{J Manag Care Spec Pharm}. Jul 2017;23(7):735–744. \\doi:10.18553/jmcp.2017.23.7.735

51. Kim KJ, Kim DH, Lee JI, Choi BK, Han IH, Nam KH. Hounsfield Units on Lumbar Computed Tomography for Predicting Regional Bone Mineral Density. \textit{Open Med (Wars)}. 2019;14:545–551. doi:10.1515/med-2019-0061

52. Pickhardt PJ, Pooler BD, Lauder T, del Rio AM, Bruce RJ, Binkley N. Opportunistic screening for osteoporosis using abdominal computed tomography scans obtained for other indications. \textit{Ann Intern Med}. Apr 16 2013;158(8):588–95. doi:10.7326/0003-4819-158-8-201304160-00003

53. Salehi S, Schlossman J, Chowdhry S, DeGaetano A, Scudeler M, Quenet S, et al. Real-World Validation of a Deep Learning AI-Based Detection Algorithm for Suspected Aortic Dissection. presented at: VEITHSymposium; 2022; New York. 

54. Administration USFaD. FDA 510(k) Clearance Letter for Viz AAA (K223443). 2023/03/17 2023;

55. Administration USFaD. FDA 510(k) Clearance Letter for BriefCase-Quantification (K230534). U.S. Department of Health and Human Services, Food and Drug Administration; 2023.

56. HeartLung C. \textit{510(k) Summary for ABMD Software (K213760)}. 2022. 2022/07/29. Accessed 2022/07/29. \href{https://www.accessdata.fda.gov/scripts/cdrh/cfdocs/cfpmn/pmn.cfm?id=K213760}{https://www.accessdata.fda.gov/scripts/cdrh/cfdocs/cfpmn/pmn.cfm?id=K213760}

57. Schreiber JJ, Anderson PA, Hsu WK. Use of computed tomography for assessing bone mineral density. \textit{Neurosurg Focus}. 2014;37(1):E4. doi:10.3171/2014.5.Focus1483

58. Ahmad A, Crawford CH, 3rd, Glassman SD, Dimar JR, 2nd, Gum JL, Carreon LY. Correlation between bone density measurements on CT or MRI versus DEXA scan: A systematic review. \textit{N Am Spine Soc J}. Jun 2023;14:100204. doi:10.1016/j.xnsj.2023.100204

59. Koo HJ, Lee J-G, Lee J-B, Kang J-W, Yang DH. Deep Learning Based Automatic Segmentation of the Thoracic Aorta from Chest Computed Tomography in Healthy Korean Adults. \textit{European Journal of Vascular and Endovascular Surgery}. 2025/01/01/ 2025;69(1):48–58. \\doi:\href{https://doi.org/10.1016/j.ejvs.2024.07.030}{https://doi.org/10.1016/j.ejvs.2024.07.030}

60. Lareyre F, Adam C, Carrier M, Dommerc C, Mialhe C, Raffort J. A fully automated pipeline for mining abdominal aortic aneurysm using image segmentation. \textit{Scientific Reports}. 2019/09/24 2019;9(1):13750. doi:10.1038/s41598-019-50251-8

61. Brutti F, Fantazzini A, Finotello A, Müller LO, Auricchio F, Pane B, et al. Deep Learning to Automatically Segment and Analyze Abdominal Aortic Aneurysm from Computed Tomography Angiography. \textit{Cardiovascular Engineering and Technology}. 2022/08/01 2022;13(4):535–547. doi:10.1007/s13239-021-00594-z

62. Caradu C, Pouncey A-L, Lakhlifi E, Brunet C, Bérard X, Ducasse E. Fully automatic volume segmentation using deep learning approaches to assess aneurysmal sac evolution after infrarenal endovascular aortic repair. \textit{Journal of Vascular Surgery}. 2022/09/01/ 2022;76(3):620–630.e3. \\
doi:\href{https://doi.org/10.1016/j.jvs.2022.03.891}{https://doi.org/10.1016/j.jvs.2022.03.891}

63. Maas EJ, Awasthi N, Pelt EGv, Sambeek MRHMv, Lopata RGP. Automatic Segmentation of Abdominal Aortic Aneurysms From Time-Resolved 3-D Ultrasound Images Using Deep Learning. \textit{IEEE Transactions on Ultrasonics, Ferroelectrics, and Frequency Control}. 2024;71(11):1420–1428. \\
doi:10.1109/TUFFC.2024.3389553

64. Abdolmanafi A, Forneris A, Moore RD, Di Martino ES. Deep-learning method for fully automatic segmentation of the abdominal aortic aneurysm from computed tomography imaging. \textit{Front Cardiovasc Med}. 2022;9:1040053. doi:10.3389/fcvm.2022.1040053

65. Roshkovan L. Opportunistic Screening: Current Challenges and Opportunities. \textit{American Journal of Roentgenology}. 2025;doi:10.2214/AJR.25.33577

66. Blankemeier L, Yao L, Long J, Reis EP, Lenchik L, Chaudhari AS, et al. Skeletal Muscle Area on CT: Determination of an Optimal Height Scaling Power and Testing for Mortality Risk Prediction. \textit{American Journal of Roentgenology}. 2024/01/01 2023;222(1):e2329889. doi:10.2214/AJR.23.29889

67. Zambrano Chaves JM, Hom J, Lenchik L, Chaudhari AS, Boutin RD. Sarcopenia, Obesity, and Sarcopenic Obesity: Retrospective Audit of Electronic Health Record Documentation versus Automated CT Analysis in 17646 Patients. \textit{Radiology}. 2025/04/01 2025;315(1):e243525. doi:10.1148/radiol.243525

68. Reis EP, Blankemeier L, Zambrano Chaves JM, Jensen MEK, Yao S, Truyts CAM, et al. Automated abdominal CT contrast phase detection using an interpretable and open-source artificial intelligence algorithm. \textit{Eur Radiol}. Oct 2024;34(10):6680–6687. doi:10.1007/s00330-024-10769-6

69. Blankemeier L, Desai A, Chaves JMZ, Wentland A, Yao S, Reis E, et al. Comp2Comp: Open-Source Body Composition Assessment on Computed Tomography. \textit{arXiv [csCV]}. 2023 

70. Ji Y, Bai H, Yang J, Ge C, Zhu Y, Zhang R, et al. AMOS: A Large-Scale Abdominal Multi-Organ Benchmark for Versatile Medical Image Segmentation. \textit{arXiv}. 2022;2206.08023

71. National Center for Health S. The 1999-2006 Dual Energy X-ray Absorptiometry (DXA) Multiple Imputation Data Files and Technical Documentation. 

72. Sarafrazi N, Wambogo EA, Shepherd JA. \textit{Osteoporosis or Low Bone Mass in Older Adults: United States, 2017–2018}. 2021. 2021/03. \href{https://www.cdc.gov/nchs/data/databriefs/db405-H.pdf}{https://www.cdc.gov/nchs/data/databriefs/db405-H.pdf}

73. Schreiber JJ, Anderson PA, Rosas HG, Buchholz AL, Au AG. Hounsfield units for assessing bone mineral density and strength: a tool for osteoporosis management. \textit{J Bone Joint Surg Am}. Jun 1 2011;93(11):1057–63.\\ doi:10.2106/jbjs.J.00160

74. Zou D, Jiang S, Zhou S, Sun Z, Zhong W, Du G, et al. Prevalence of Osteoporosis in Patients Undergoing Lumbar Fusion for Lumbar Degenerative Diseases: A Combination of DXA and Hounsfield Units. \textit{Spine (Phila Pa 1976)}. Apr 1 2020;45(7):E406–e410. \\doi:10.1097/brs.0000000000003284

\section{\textbf{Appendix:}}

\textbf{Appendix 1:} Confusion matrices of BMD performance by subgroup with percentage as proportion correctly/incorrectly categorized for a given ground truth value, and associated sensitivity/specificities.

\small
\setlength{\tabcolsep}{4pt}
\renewcommand{\arraystretch}{0.95}

\begin{longtable}{@{}
>{\raggedright\arraybackslash}p{0.32\linewidth}
>{\raggedright\arraybackslash}p{0.15\linewidth}
>{\raggedright\arraybackslash}p{0.15\linewidth}
>{\raggedright\arraybackslash}p{0.32\linewidth}
@{}}

\toprule
\textbf{Age (years)} &
\multicolumn{2}{c}{\textbf{DXA T-score: Normal BMD}} &
\textbf{DXA T-score: Low BMD} \\
\cmidrule(lr){2-3}\cmidrule(lr){4-4}
\endfirsthead

\toprule
\textbf{Age (years)} &
\multicolumn{2}{c}{\textbf{DXA T-score: Normal BMD}} &
\textbf{DXA T-score: Low BMD} \\
\cmidrule(lr){2-3}\cmidrule(lr){4-4}
\endhead

18--21 (N = 1) & \multicolumn{2}{c}{N = 0} & N = 1 \\
Model: Normal BMD & \multicolumn{2}{c}{0 (0\%)} & 0 (0\%) \\
Model: Low BMD    & \multicolumn{2}{c}{0 (0\%)} & 1 (100\%) \\
\multicolumn{4}{@{}l@{}}{Sensitivity: 100\%\hspace{1.5em}Specificity: N/A} \\
\midrule

22--29 (N = 5) & \multicolumn{2}{c}{N = 3} & N = 2 \\
Model: Normal BMD & \multicolumn{2}{c}{3 (100\%)} & 1 (50\%) \\
Model: Low BMD    & \multicolumn{2}{c}{0 (0\%)}   & 1 (50\%) \\
\multicolumn{4}{@{}l@{}}{Sensitivity: 50\%\hspace{1.5em}Specificity: 100\%} \\
\midrule

30--39 (N = 15) & \multicolumn{2}{c}{N = 9} & N = 6 \\
Model: Normal BMD & \multicolumn{2}{c}{9 (100\%)} & 3 (50\%) \\
Model: Low BMD    & \multicolumn{2}{c}{0 (0\%)}   & 3 (50\%) \\
\multicolumn{4}{@{}l@{}}{Sensitivity: 50\%\hspace{1.5em}Specificity: 100\%} \\
\midrule

40--49 (N = 29) & \multicolumn{2}{c}{N = 20} & N = 9 \\
Model: Normal BMD & \multicolumn{2}{c}{19 (95\%)} & 3 (33\%) \\
Model: Low BMD    & \multicolumn{2}{c}{1 (5\%)}   & 6 (67\%) \\
\multicolumn{4}{@{}l@{}}{Sensitivity: 67\%\hspace{1.5em}Specificity: 95\%} \\
\midrule

50--69 (N = 211) & \multicolumn{2}{c}{N = 123} & N = 88 \\
Model: Normal BMD & \multicolumn{2}{c}{100 (81\%)} & 18 (20\%) \\
Model: Low BMD    & \multicolumn{2}{c}{23 (19\%)}  & 70 (80\%) \\
\multicolumn{4}{@{}l@{}}{Sensitivity: 80\%\hspace{1.5em}Specificity: 81\%} \\
\midrule

70+ (N = 110) & \multicolumn{2}{c}{N = 58} & N = 52 \\
Model: Normal BMD & \multicolumn{2}{c}{36 (62\%)} & 5 (10\%) \\
Model: Low BMD    & \multicolumn{2}{c}{22 (38\%)} & 47 (90\%) \\
\multicolumn{4}{@{}l@{}}{Sensitivity: 90\%\hspace{1.5em}Specificity: 62\%} \\

\bottomrule
\end{longtable}

\small
\setlength{\tabcolsep}{4pt}
\renewcommand{\arraystretch}{0.95}

\begin{longtable}{@{}
>{\raggedright\arraybackslash}p{0.32\linewidth}
>{\raggedright\arraybackslash}p{0.15\linewidth}
>{\raggedright\arraybackslash}p{0.15\linewidth}
>{\raggedright\arraybackslash}p{0.32\linewidth}
@{}}

\toprule
\textbf{Sex} &
\multicolumn{2}{c}{\textbf{DXA T-score: Normal BMD}} &
\textbf{DXA T-score: Low BMD} \\
\cmidrule(lr){2-3}\cmidrule(lr){4-4}
\endfirsthead

\toprule
\textbf{Sex} &
\multicolumn{2}{c}{\textbf{DXA T-score: Normal BMD}} &
\textbf{DXA T-score: Low BMD} \\
\cmidrule(lr){2-3}\cmidrule(lr){4-4}
\endhead

Male (n = 63) & \multicolumn{2}{c}{N = 45} & N = 18 \\
Model: Normal BMD & \multicolumn{2}{c}{34 (76\%)} & 1 (6\%) \\
Model: Low BMD    & \multicolumn{2}{c}{11 (24\%)} & 17 (94\%) \\
\multicolumn{4}{@{}l@{}}{Sensitivity: 94\%\hspace{1.5em}Specificity: 76\%} \\
\midrule

Female (n = 308) & \multicolumn{2}{c}{N = 168} & N = 140 \\
Model: Normal BMD & \multicolumn{2}{c}{133 (79\%)} & 29 (21\%) \\
Model: Low BMD    & \multicolumn{2}{c}{35 (21\%)}  & 111 (79\%) \\
\multicolumn{4}{@{}l@{}}{Sensitivity: 79\%\hspace{1.5em}Specificity: 79\%} \\

\bottomrule
\end{longtable}

\small
\setlength{\tabcolsep}{4pt}
\renewcommand{\arraystretch}{0.95}

\begin{longtable}{@{}
>{\raggedright\arraybackslash}p{0.32\linewidth}
>{\raggedright\arraybackslash}p{0.15\linewidth}
>{\raggedright\arraybackslash}p{0.15\linewidth}
>{\raggedright\arraybackslash}p{0.32\linewidth}
@{}}

\toprule
\textbf{Manufacturer} &
\multicolumn{2}{c}{\textbf{DXA T-score: Normal BMD}} &
\textbf{DXA T-score: Low BMD} \\
\cmidrule(lr){2-3}\cmidrule(lr){4-4}
\endfirsthead

\toprule
\textbf{Manufacturer} &
\multicolumn{2}{c}{\textbf{DXA T-score: Normal BMD}} &
\textbf{DXA T-score: Low BMD} \\
\cmidrule(lr){2-3}\cmidrule(lr){4-4}
\endhead

GE (n = 122) & \multicolumn{2}{c}{N = 77} & N = 45 \\
Model: Normal BMD & \multicolumn{2}{c}{61 (87\%)} & 10 (22\%) \\
Model: Low BMD    & \multicolumn{2}{c}{16 (13\%)} & 35 (78\%) \\
\multicolumn{4}{@{}l@{}}{Sensitivity: 78\%\hspace{1.5em}Specificity: 87\%} \\
\midrule

Philips (n = 151) & \multicolumn{2}{c}{N = 78} & N = 73 \\
Model: Normal BMD & \multicolumn{2}{c}{54 (69\%)} & 13 (18\%) \\
Model: Low BMD    & \multicolumn{2}{c}{24 (31\%)} & 60 (82\%) \\
\multicolumn{4}{@{}l@{}}{Sensitivity: 82\%\hspace{1.5em}Specificity: 69\%} \\
\midrule

Siemens (n = 66) & \multicolumn{2}{c}{N = 32} & N = 34 \\
Model: Normal BMD & \multicolumn{2}{c}{28 (88\%)} & 6 (18\%) \\
Model: Low BMD    & \multicolumn{2}{c}{4 (12\%)}  & 28 (82\%) \\
\multicolumn{4}{@{}l@{}}{Sensitivity: 82\%\hspace{1.5em}Specificity: 88\%} \\
\midrule

Toshiba (n = 32) & \multicolumn{2}{c}{N = 26} & N = 6 \\
Model: Normal BMD & \multicolumn{2}{c}{24 (92\%)} & 1 (17\%) \\
Model: Low BMD    & \multicolumn{2}{c}{2 (8\%)}  & 5 (83\%) \\
\multicolumn{4}{@{}l@{}}{Sensitivity: 83\%\hspace{1.5em}Specificity: 92\%} \\

\bottomrule
\end{longtable}

\small
\setlength{\tabcolsep}{4pt}
\renewcommand{\arraystretch}{0.95}

\begin{longtable}{@{}
>{\raggedright\arraybackslash}p{0.32\linewidth}
>{\raggedright\arraybackslash}p{0.15\linewidth}
>{\raggedright\arraybackslash}p{0.15\linewidth}
>{\raggedright\arraybackslash}p{0.32\linewidth}
@{}}

\toprule
\textbf{Slice thickness} &
\multicolumn{2}{c}{\textbf{DXA T-score: Normal BMD}} &
\textbf{DXA T-score: Low BMD} \\
\cmidrule(lr){2-3}\cmidrule(lr){4-4}
\endfirsthead

\toprule
\textbf{Slice thickness} &
\multicolumn{2}{c}{\textbf{DXA T-score: Normal BMD}} &
\textbf{DXA T-score: Low BMD} \\
\cmidrule(lr){2-3}\cmidrule(lr){4-4}
\endhead

Slice $<$ 2.0 mm (n = 77) & \multicolumn{2}{c}{N = 58} & N = 19 \\
Model: Normal BMD & \multicolumn{2}{c}{49 (84\%)} & 2 (11\%) \\
Model: Low BMD    & \multicolumn{2}{c}{9 (16\%)}  & 17 (89\%) \\
\multicolumn{4}{@{}l@{}}{Sensitivity: 89\%\hspace{1.5em}Specificity: 84\%} \\
\midrule

Slice 2.0--2.5 mm (n = 68) & \multicolumn{2}{c}{N = 35} & N = 33 \\
Model: Normal BMD & \multicolumn{2}{c}{29 (83\%)} & 10 (30\%) \\
Model: Low BMD    & \multicolumn{2}{c}{6 (17\%)}  & 23 (70\%) \\
\multicolumn{4}{@{}l@{}}{Sensitivity: 70\%\hspace{1.5em}Specificity: 83\%} \\
\midrule

Slice 3.0 mm (n = 169) & \multicolumn{2}{c}{N = 84} & N = 85 \\
Model: Normal BMD & \multicolumn{2}{c}{65 (77\%)} & 16 (19\%) \\
Model: Low BMD    & \multicolumn{2}{c}{19 (23\%)} & 69 (81\%) \\
\multicolumn{4}{@{}l@{}}{Sensitivity: 81\%\hspace{1.5em}Specificity: 77\%} \\
\midrule

Slice $>$ 3.0 mm (n = 57) & \multicolumn{2}{c}{N = 36} & N = 21 \\
Model: Normal BMD & \multicolumn{2}{c}{24 (67\%)} & 2 (10\%) \\
Model: Low BMD    & \multicolumn{2}{c}{12 (33\%)} & 19 (90\%) \\
\multicolumn{4}{@{}l@{}}{Sensitivity: 90\%\hspace{1.5em}Specificity: 67\%} \\

\bottomrule
\end{longtable}

\small
\setlength{\tabcolsep}{4pt}
\renewcommand{\arraystretch}{0.95}

\begin{longtable}{@{}
>{\raggedright\arraybackslash}p{0.32\linewidth}
>{\raggedright\arraybackslash}p{0.15\linewidth}
>{\raggedright\arraybackslash}p{0.15\linewidth}
>{\raggedright\arraybackslash}p{0.32\linewidth}
@{}}

\toprule
\textbf{Reconstruction kernel} &
\multicolumn{2}{c}{\textbf{DXA T-score: Normal BMD}} &
\textbf{DXA T-score: Low BMD} \\
\cmidrule(lr){2-3}\cmidrule(lr){4-4}
\endfirsthead

\toprule
\textbf{Reconstruction kernel} &
\multicolumn{2}{c}{\textbf{DXA T-score: Normal BMD}} &
\textbf{DXA T-score: Low BMD} \\
\cmidrule(lr){2-3}\cmidrule(lr){4-4}
\endhead

GE: Standard (n = 122) & \multicolumn{2}{c}{N = 77} & N = 45 \\
Model: Normal BMD & \multicolumn{2}{c}{61 (79\%)} & 10 (22\%) \\
Model: Low BMD    & \multicolumn{2}{c}{16 (21\%)} & 35 (78\%) \\
\multicolumn{4}{@{}l@{}}{Sensitivity: 78\%\hspace{1.5em}Specificity: 79\%} \\
\midrule

Philips: B (n = 131) & \multicolumn{2}{c}{N = 68} & N = 63 \\
Model: Normal BMD & \multicolumn{2}{c}{46 (68\%)} & 13 (21\%) \\
Model: Low BMD    & \multicolumn{2}{c}{22 (32\%)} & 50 (79\%) \\
\multicolumn{4}{@{}l@{}}{Sensitivity: 79\%\hspace{1.5em}Specificity: 68\%} \\
\midrule

Philips: Other (n = 20) & \multicolumn{2}{c}{N = 10} & N = 10 \\
Model: Normal BMD & \multicolumn{2}{c}{8 (80\%)} & 0 (0\%) \\
Model: Low BMD    & \multicolumn{2}{c}{2 (20\%)} & 10 (100\%) \\
\multicolumn{4}{@{}l@{}}{Sensitivity: 100\%\hspace{1.5em}Specificity: 80\%} \\
\midrule

Siemens: B30f (n = 21) & \multicolumn{2}{c}{N = 11} & N = 10 \\
Model: Normal BMD & \multicolumn{2}{c}{9 (82\%)} & 3 (30\%) \\
Model: Low BMD    & \multicolumn{2}{c}{2 (18\%)} & 7 (70\%) \\
\multicolumn{4}{@{}l@{}}{Sensitivity: 70\%\hspace{1.5em}Specificity: 82\%} \\
\midrule

Siemens: Other (n = 45) & \multicolumn{2}{c}{N = 21} & N = 24 \\
Model: Normal BMD & \multicolumn{2}{c}{19 (90\%)} & 3 (13\%) \\
Model: Low BMD    & \multicolumn{2}{c}{2 (10\%)} & 21 (87\%) \\
\multicolumn{4}{@{}l@{}}{Sensitivity: 87\%\hspace{1.5em}Specificity: 90\%} \\
\midrule

Toshiba: FC08 (n = 19) & \multicolumn{2}{c}{N = 15} & N = 4 \\
Model: Normal BMD & \multicolumn{2}{c}{14 (93\%)} & 0 (0\%) \\
Model: Low BMD    & \multicolumn{2}{c}{1 (7\%)} & 4 (100\%) \\
\multicolumn{4}{@{}l@{}}{Sensitivity: 100\%\hspace{1.5em}Specificity: 93\%} \\
\midrule

Toshiba: Other (n = 13) & \multicolumn{2}{c}{N = 11} & N = 2 \\
Model: Normal BMD & \multicolumn{2}{c}{10 (91\%)} & 1 (50\%) \\
Model: Low BMD    & \multicolumn{2}{c}{1 (9\%)} & 1 (50\%) \\
\multicolumn{4}{@{}l@{}}{Sensitivity: 50\%\hspace{1.5em}Specificity: 91\%} \\

\bottomrule
\end{longtable}

\small
\setlength{\tabcolsep}{4pt}
\renewcommand{\arraystretch}{0.95}

\begin{longtable}{@{}
>{\raggedright\arraybackslash}p{0.32\linewidth}
>{\raggedright\arraybackslash}p{0.15\linewidth}
>{\raggedright\arraybackslash}p{0.15\linewidth}
>{\raggedright\arraybackslash}p{0.32\linewidth}
@{}}

\toprule
\textbf{Site} &
\multicolumn{2}{c}{\textbf{DXA T-score: Normal BMD}} &
\textbf{DXA T-score: Low BMD} \\
\cmidrule(lr){2-3}\cmidrule(lr){4-4}
\endfirsthead

\toprule
\textbf{Site} &
\multicolumn{2}{c}{\textbf{DXA T-score: Normal BMD}} &
\textbf{DXA T-score: Low BMD} \\
\cmidrule(lr){2-3}\cmidrule(lr){4-4}
\endhead

MedStar (n = 45) & \multicolumn{2}{c}{N = 30} & N = 15 \\
Model: Normal BMD & \multicolumn{2}{c}{25 (17\%)} & 0 (0\%) \\
Model: Low BMD    & \multicolumn{2}{c}{5 (83\%)}  & 15 (100\%) \\
\multicolumn{4}{@{}l@{}}{Sensitivity: 100\%\hspace{1.5em}Specificity: 83\%} \\
\midrule

DASA (n = 94) & \multicolumn{2}{c}{N = 69} & N = 25 \\
Model: Normal BMD & \multicolumn{2}{c}{58 (84\%)} & 2 (8\%) \\
Model: Low BMD    & \multicolumn{2}{c}{11 (16\%)} & 23 (92\%) \\
\multicolumn{4}{@{}l@{}}{Sensitivity: 92\%\hspace{1.5em}Specificity: 84\%} \\
\midrule

UH (n = 139) & \multicolumn{2}{c}{N = 70} & N = 69 \\
Model: Normal BMD & \multicolumn{2}{c}{51 (73\%)} & 16 (23\%) \\
Model: Low BMD    & \multicolumn{2}{c}{19 (27\%)} & 53 (77\%) \\
\multicolumn{4}{@{}l@{}}{Sensitivity: 77\%\hspace{1.5em}Specificity: 73\%} \\
\midrule

UAB (n = 93) & \multicolumn{2}{c}{N = 44} & N = 49 \\
Model: Normal BMD & \multicolumn{2}{c}{33 (75\%)} & 12 (24\%) \\
Model: Low BMD    & \multicolumn{2}{c}{11 (25\%)} & 37 (76\%) \\
\multicolumn{4}{@{}l@{}}{Sensitivity: 76\%\hspace{1.5em}Specificity: 75\%} \\

\bottomrule
\end{longtable}

\FloatBarrier
 \textbf{Appendix 2:} Allowed CT reconstruction kernels in the BMD pipeline.

GE MEDICAL SYSTEMS

- STANDARD

 Philips

- B

- C

- SB

 SIEMENS

- B20f

- B30f

- B31f

- B40f

- Bf37f

- Br38f

- Br40d

- I30f

- I40f

 TOSHIBA

- FC02

- FC07

- FC08

\end{document}